\documentclass[10pt,twocolumn,letterpaper]{article}

\usepackage[pagenumbers]{wacv} 

\usepackage{colortbl}
\usepackage[inkscapelatex=false]{svg}
\usepackage{graphicx} 
\usepackage{caption}
\usepackage{arydshln}
\usepackage{float}
\usepackage{tabularx}
\usepackage[table]{xcolor}
\definecolor{firstc}{HTML}{C0E2CA} 
\definecolor{secondc}{HTML}{E2EDB9} 
\definecolor{thirdc}{HTML}{FFFAC1} 

\definecolor{wacvblue}{rgb}{0.21,0.49,0.74}
\usepackage[pagebackref,breaklinks,colorlinks,allcolors=wacvblue]{hyperref}

\def\wacvPaperID{2628} 
\def\confName{WACV}
\def\confYear{2027}

\title{SCOUT-SLAM: Structurally-Coupled Dual Uncertainty-Aware 3DGS SLAM in the Wild}

\author{Kumaran Karthik\\
\and
Pramat Shastri Jois\textsuperscript{1}\\
\textsuperscript{1}Indian Institute of Science\\
\and
Suresh Sundaram\textsuperscript{1}\\
\\
}
\usepackage{capt-of}

\begin{document}
\twocolumn[{%
\renewcommand\twocolumn[1][]{#1}%
\maketitle
}]
\begin{abstract}
Recently, 3D Gaussian Splatting SLAM (3DGS-SLAM) has gained significant momentum in simultaneous localization and 3DGS scene reconstruction. In real-world scenarios with rapid camera motion and cluttered dynamic environments, existing methods rely on the stability of the underlying scene reconstruction to model uncertainty. This leads to a \textit{circular dependency} between camera tracking accuracy and reconstruction quality: reconstruction instabilities degrade uncertainty modeling, which affects accurate camera tracking and static scene reconstruction. To address this, the paper proposes \textbf{SCOUT-SLAM}, a structurally-coupled dual-uncertainty framework in which both uncertainties are estimated from a shared base network. A low-rank adaptation of this network, trained on multi-view feature consistency, estimates a tracking uncertainty that does not depend solely on the reconstruction quality. A spatially-adaptive prior modulates the network's training objective so that reconstruction instability does not inflate uncertainty on static regions, keeping the shared representation intact for both branches. Evaluations on dynamic benchmarks (TUM RGB-D, Bonn Dynamic, Wild-SLAM MoCap) demonstrate that SCOUT-SLAM achieves state-of-the-art camera tracking accuracy and artifact-free static scene reconstruction. Code is available at: \url{https://github.com/kumaran-3527/SCOUT-SLAM}
\end{abstract}
    
\begin{figure*}[t]
    \centering
    \includegraphics[width=0.85\textwidth]{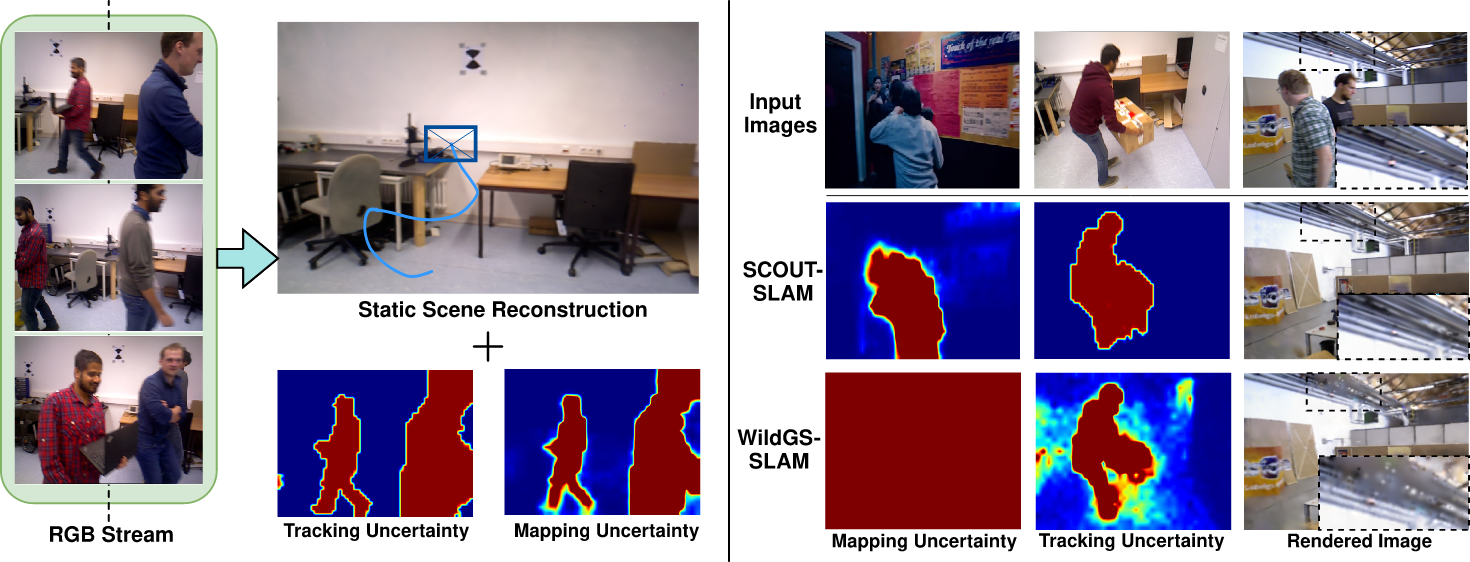}
    \captionof{figure}{\textbf{SCOUT-SLAM.} \textit{Left}: Given a captured \textit{in-the-Wild} video, SCOUT-SLAM estimates, camera trajectory, high-fidelity static scene and uncertainties. \textit{Right}: SCOUT-SLAM's uncertainty estimates tightly isolates dynamic objects to preserve high frequency static geometry.}
    \label{fig:teaser}
\end{figure*}
\section{Introduction}
\label{sec:intro}
Visual Simultaneous Localization and Mapping (SLAM) jointly tracks the camera motion and builds scene representations from sequential imagery~\cite{mur2017orb, engel2017direct, teed2021droid}. SLAM has widespread application in autonomous navigation~\cite{gaussnav2025,geiger2013vision,liu2025slideslam}, augmented reality~\cite{splatloc}, and robotic manipulation~\cite{hoque2026egodex}. With the advent of 3D Gaussian Splatting (3DGS)~\cite{kerbl20233d}, 3DGS-SLAM pipelines can now produce high-fidelity scene reconstructions~\cite{Matsuki_2024_CVPR, keetha2024splatam, Sandstrom_2025_CVPR, yan2023gs}. High-fidelity reconstructions enable downstream robotic tasks such as end-to-end visual navigation~\cite{gaussnav2025, splatnav2025, chen2025grad}, semantic scene understanding~\cite{nisslam2024}, and goal-conditioned planning~\cite{beings2024, iglnav2025}. Despite significant progress, achieving reliable 3DGS-SLAM in real-world environments is challenging. Dynamic and non-rigid objects often degrade pose estimation and 3DGS reconstruction, limiting the practicality of 3DGS-SLAM systems.

To handle dynamic environments, recent neural~\cite{jiang2024rodyn, li2025ddn, schischka2024dynamon} and 3DGS~\cite{xu2024dg,li2025dy3dgs} SLAM approaches derive dynamic masks by fusing motion cues with semantic or depth priors. However, they heavily depend on predefined semantic categories of dynamic objects, which limits applicability in real-world environments. Recent uncertainty-aware 3DGS-SLAM systems~\cite{zheng2025wildgs,zheng2026upslam} adapt the uncertainty modeling developed for offline novel-view synthesis~\cite{ren2024nerf,kulhanek2024wildgaussians} to mask out dynamic objects and transient elements without predefined semantic categories.

However, these approaches derive uncertainty from the underlying scene reconstruction, which introduces a \textit{circular dependency} between camera tracking accuracy and reconstruction quality. During fast camera motion in cluttered dynamic environments, the 3DGS reconstruction becomes unstable, degrading the uncertainty estimates that camera tracking relies on, which further corrupts the reconstruction. This results in inaccurate camera tracking and scene reconstruction in such scenarios.

To address this \textit{circular dependency}, we propose \textbf{SCOUT-SLAM} (\textbf{S}tructurally-\textbf{CO}upled dual-\textbf{U}ncertain\textbf{T}y SLAM), a monocular 3DGS-SLAM framework that maintains structurally coupled estimates of tracking and mapping uncertainty. We term the framework \textit{structurally coupled} because the two uncertainties share a single base network, the Mapping Uncertainty Network (\textbf{MUNet}), a shallow MLP that predicts a per-pixel uncertainty from DINOv2~\cite{oquab2023dinov2,yue2024improving}, trained against the residual between input and rendered views. The tracking uncertainty is estimated by a Tracking Uncertainty Adapter (\textbf{TUA}), a low-rank adaptation (LoRA)~\cite{hu2022lora} of MUNet optimized at each Bundle Adjustment (BA) iteration on a Multi-view Feature Consistency (MFC) loss over DINOv2~\cite{oquab2023dinov2,yue2024improving} features. The MFC loss is computed directly between observed views. The TUA therefore estimates a tracking uncertainty that does not depend solely on the current reconstruction quality. This uncertainty then reweights the residuals used in pose optimization. Complementing this, a spatially-adaptive prior modulates the per-pixel log-likelihood term of the Negative Log-Likelihood (NLL) objective used to train MUNet, penalizing corrupted estimates so that reconstruction instability does not inflate uncertainty on static geometry. Since the TUA is a low-rank adaptation of MUNet, it inherits the latter's feature space, protecting MUNet from degradation therefore serves both branches: the prior keeps the shared representation stable allowing the TUA to estimate reliable tracking uncertainty. Together, these components yield stable uncertainties under fast camera motion in cluttered, dynamic environments. Our contributions are:
\begin{itemize}
    \item \textbf{SCOUT-SLAM}, a structurally coupled dual-uncertainty framework for monocular 3DGS-SLAM. Tracking and mapping uncertainty are estimated from a single shared base network, so each depends on the other remaining well-conditioned, breaking the \textit{circular dependency} between camera tracking accuracy and reconstruction quality under fast camera motion in cluttered dynamic environments.

    \item \textbf{Tracking Uncertainty Adapter (TUA)}, a low-rank adaptation of MUNet trained on multi-view feature consistency. It reweights the residuals in pose optimization with a tracking uncertainty that does not depend solely on the current reconstruction quality, while inheriting the shared feature space it adapts.

    \item A \textbf{spatially-adaptive prior} that modulates the per-pixel log-likelihood in the NLL objective of MUNet. It prevents reconstruction instability from inflating uncertainty on static geometry, keeping the shared base weights uncorrupted for both branches and yielding static 3DGS reconstructions.

    \item SCOUT-SLAM is evaluated on three dynamic benchmarks (TUM RGB-D, Bonn Dynamic, Wild-SLAM MoCap) against baselines demonstrating state-of-the-art tracking accuracy and consistent improvements in static reconstruction fidelity.
\end{itemize}

\begin{figure*}[t]
    \centering
    \includegraphics[width=0.95\textwidth]{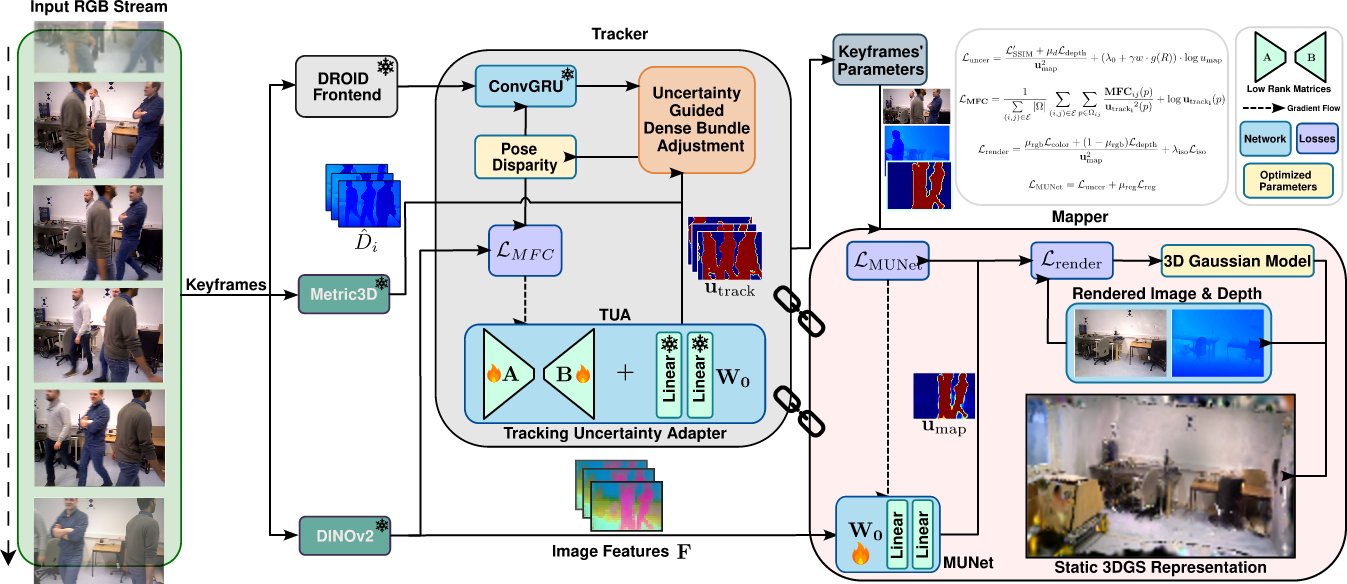}
    \caption{\textbf{System Overview.} SCOUT-SLAM simultaneously tracks camera motion while reconstructing a 3DGS static scene. The structurally-coupled dual uncertainty framework addresses the circular dependency between camera tracking accuracy and reconstruction quality that arises when uncertainty modeling is tightly coupled with the reconstruction quality. From an input RGB stream, DINOv2 features $\mathbf{F}$ are extracted and decoded through a shared shallow MLP, the mapping uncertainty network (MUNet). \textbf{Tracker:} A Tracking Uncertainty Adapter (TUA), a low-rank adaptation over the MUNet, is optimized on multi-view feature consistency (MFC) loss $\mathcal{L}_\text{MFC}$. Consequently, tracking uncertainty estimation $u_{track}$ is not solely dependent on reconstruction quality. The resulting tracking confidence mask $W$ down-weights dynamic objects during uncertainty-guided dense bundle adjustment (DBA). \textbf{Mapper:} A spatially-adaptive prior modulates the MUNet's training objective to prevent uncertainty inflation on static regions ensuring 3DGS reconstruction stability keeping the shared feature space intact. The per-pixel uncertainty $u_{map}$ is used for uncertainty-aware 3DGS mapping resulting in a artifact-free 3DGS static scene. Together, these components address the circular dependency between camera tracking accuracy and reconstruction quality during fast camera motion in cluttered dynamic environments.}
    
    \label{fig:system_overview}
    
\end{figure*}
\section{Related Works}
\label{sec:related-works}

\subsection{Neural and GS-based SLAM}
Neural implicit representations introduced a class of dense SLAM systems where scene geometry and appearance are encoded in a neural network and jointly optimized with camera poses. iMAP~\cite{sucar2021imap} and NICE-SLAM~\cite{zhu2022nice} were the first to demonstrate this paradigm on RGB-D sequences, with later works such as Co-SLAM~\cite{wang2023coslam} and ESLAM~\cite{johari2023eslam} improving scalability through hybrid encodings. In particular, to handle dynamic scenes, NeRF-based systems such as RoDyn-SLAM~\cite{jiang2024rodyn} and DynaMon~\cite{schischka2024dynamon} combine optical flow with semantic priors to filter out transient regions. However, high computational overhead and sub-real-time rendering severely constrain their online applicability.

The emergence of 3D Gaussian Splatting (3DGS)~\cite{kerbl20233d} provided an explicit, rasterization-based representation that accelerates both rendering and map updates, enabling SLAM systems to run at interactive rates. MonoGS~\cite{Matsuki_2024_CVPR} was the first near-real-time monocular Gaussian Splatting (GS) SLAM to derive pose gradients for joint pose-scene optimization. Methods such as GS-SLAM~\cite{yan2023gs} perform tracking over a sparsely selected set of Gaussians, Photo-SLAM~\cite{huang2024photo} couples ORB-SLAM3 tracking with 3DGS mapping, and SplaTAM~\cite{keetha2024splatam} guides online tracking and mapping via a silhouette mask. Among these, Splat-SLAM~\cite{Sandstrom_2025_CVPR} is a leading monocular Gaussian system that extends its design to enable globally consistent mapping. The photometric and geometric consistency these systems rely on is violated when moving objects are present, since the scene is assumed to be predominantly static.

To handle dynamic objects, DG-SLAM~\cite{xu2024dg}, DDN-SLAM~\cite{li2025ddn} and Dy3DGS-SLAM~\cite{li2025dy3dgs} leverage semantic or motion cues, but their reliance on closed-set categories limits generalization. While UP-SLAM~\cite{zheng2026upslam} computes uncertainty from rendering residuals to achieve open-set masking, it still requires a well-initialized static scene reconstruction, necessitating closed-set priors (e.g., YOLOv8-seg) during initialization. WildGS-SLAM~\cite{zheng2025wildgs} avoids semantic priors by decoding DINOv2~\cite{oquab2023dinov2,yue2024improving} features into learned pixel uncertainties for joint tracking and mapping, inherently coupling tracking accuracy to reconstruction fidelity. In contrast, the proposed method estimates a dedicated tracking uncertainty directly from multi-view feature consistency, thereby enabling native open-set dynamic handling from initialization without semantic priors or map-quality bottlenecks.

\subsection{Traditional SLAM}
Traditional visual SLAM tracks camera motion directly via sparse feature correspondences~\cite{klein2007parallel,campos2021orbslam3} or direct photometric alignment~\cite{engel2014lsd,engel2017direct}. Such systems assume a static scene and drift in the presence of moving objects due to improper correspondences.Dynamic variants tackle this through residual analysis (StaticFusion~\cite{scona2018staticfusion},
ReFusion~\cite{refusion3d}) or by masking known object classes (DynaSLAM~\cite{bescos2018dynaslam},
DS-SLAM~\cite{yu2018ds}). DROID-SLAM \cite{teed2021droid} recasts correspondence-based tracking in a learned form, training a recurrent update operator that drives a dense, differentiable bundle adjustment (DBA) layer to jointly refine poses and geometry, with DPVO \cite{teed2023deep} following a lighter, patch-based approach. To handle dynamic motion, MegaSaM~\cite{li2025megasam} integrates recurrent flow-state motion predictions, though large-scale synthetic training limits its real-world generalization. WildPose~\cite{zheng2026wildpose} couples a frozen 3D foundation backbone (MASt3R~\cite{mast3r_eccv24}) with a retrained motion-mask detector. DROID-W \cite{li2026droid} folds per-pixel uncertainty directly into the DBA layer, estimated from multi-view feature consistency rather than the reconstruction, yielding robust tracking under unknown dynamics without relying on a map. However, decoupling tracking from the map entirely discards the reconstruction priors that could otherwise reinforce it.

\section{Structurally-Coupled Dual Uncertainty Framework}
\label{sec:method}
Given a monocular RGB stream in a dynamic environment, SCOUT-SLAM tracks the camera motion and reconstructs a static 3DGS scene. \cref{fig:system_overview} shows the system overview. DINOv2~\cite{oquab2023dinov2, yue2024improving} features $\mathbf{F}_i$ of each keyframe are extracted and decoded by a single shared network, named the Mapping Uncertainty Network (MUNet). A Tracking Uncertainty Adapter (TUA) computes per-pixel tracking uncertainty used in pose optimization that reweights the residuals in pose optimization. The TUA is a low-rank adaptation (LoRA)~\cite{hu2022lora} over the MUNet. The low-rank matrices are optimized on multi-view feature consistency at each bundle adjustment iteration, so camera tracking accuracy is no longer tightly coupled to the reconstruction quality. A spatially-adaptive prior modulates the MUNet's training objective to prevent uncertainty inflation on static regions ensuring 3DGS reconstruction stability keeping the shared feature space intact. Together, these components address the circular dependency between camera tracking accuracy and reconstruction quality during fast camera motion in cluttered dynamic environments.
\subsection{Preliminaries: 3D Gaussian Splatting}
\label{subsec:prelim_3dgs}
3D Gaussian Splatting (3DGS)~\cite{kerbl20233d} represents a scene explicitly using a set of 3D Gaussians, each parameterized by a center $\mu \in \mathbb{R}^3$, scaling $S$, rotation $R$, opacity $\alpha \in [0, 1]$, and Spherical Harmonics (SH) coefficients. Pixel color $\hat{C}$ and depth $\hat{D}$ are computed by tile-based rasterization, depth-sorting, and $\alpha$-blending:
{\small
\begin{equation}
\hat{C} = \sum_{i \in \mathcal{N}} c_i \alpha_i \prod_{j=1}^{i-1} (1 - \alpha_j), \quad \hat{D} = \sum_{i \in \mathcal{N}} d_i \alpha_i \prod_{j=1}^{i-1} (1 - \alpha_j),
\end{equation}
}%
where $c_i$ and $d_i$ are the color and depth of the $i$-the Gaussian, and $\mathcal{N}$ is the ordered set of overlapping Gaussians. The explicit representation allows for efficient differentiable rasterization and real-time rendering. In proposed method, the gaussian parameters are optimized using mapping uncertainty to reconstruct the static background.
\subsection{Preliminaries: Low-Rank Adaptation (LoRA)}
\label{subsec:prelim_lora}
Low-Rank Adaptation (LoRA) \cite{hu2022lora} is a parameter-efficient method to adapt pre-trained neural networks. Instead of updating a full pre-trained weight matrix $\mathbf{W}_0 \in \mathbb{R}^{d \times k}$, LoRA freezes $\mathbf{W}_0$ and injects trainable low-rank decomposition matrices $\mathbf{B} \in \mathbb{R}^{d \times r}$ and $\mathbf{A} \in \mathbb{R}^{r \times k}$. For an input vector $\mathbf{x} \in \mathbb{R}^k$, the forward pass is computed as:
{\small
\begin{equation}
\label{eq:prelim_lora}
h = \mathbf{W}_0 \mathbf{x} + \Delta\mathbf{W} \mathbf{x} = \mathbf{W}_0 \mathbf{x} + \mathbf{B} \mathbf{A} \mathbf{x},
\end{equation}
}%
where the rank $r \ll \min(d,k)$. This isolates task-specific updates into the lightweight $\Delta\mathbf{W}$ while preserving the foundational knowledge structurally embedded in $\mathbf{W}_0$. TUA is a low-rank adaptation of the MUNet on the multi-view feature consistency across BA window.

\begin{figure*}[t]
    \centering
    \setlength{\tabcolsep}{1pt}
    \renewcommand{\arraystretch}{0.5}
    %
    \newcommand{\cw}{0.15\textwidth}
    \newcommand{\ch}{0.080\textwidth}
    \begin{tabular}{cccccc}
        {\scriptsize Input} &
        {\scriptsize WildGS-SLAM~\cite{zheng2025wildgs}} &
        {\scriptsize DROID-W~\cite{li2026droid}} &
        {\scriptsize MegaSaM~\cite{li2025megasam}} &
        {\scriptsize Tracking uncer. (Ours)} &
        {\scriptsize Mapping uncer. (Ours)} \\[2pt]
        \includegraphics[width=\cw, height=\ch]{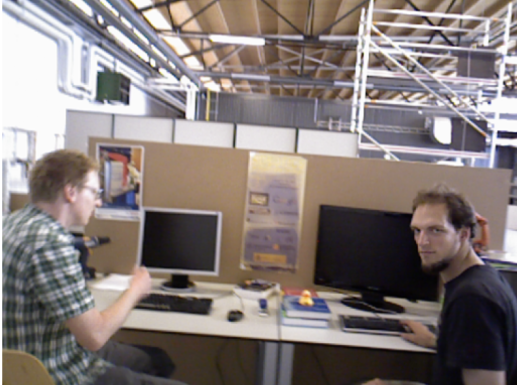} &
        \includegraphics[width=\cw, height=\ch]{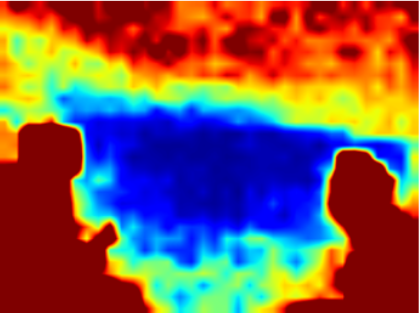} &
        \includegraphics[width=\cw, height=\ch]{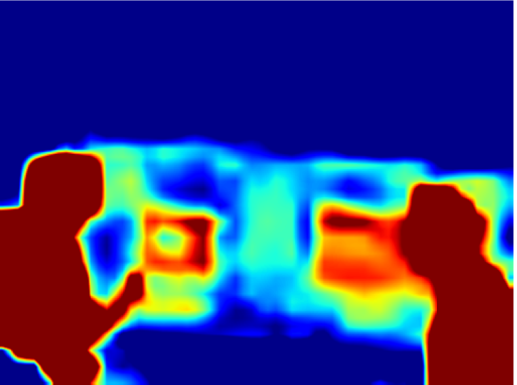} &
        \includegraphics[width=\cw, height=\ch]{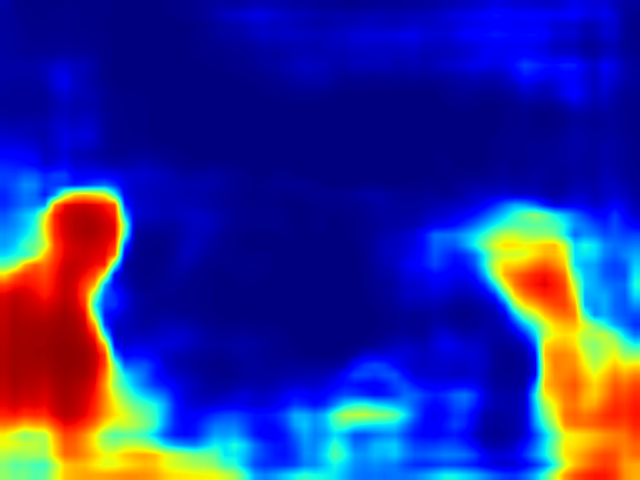} &
        \includegraphics[width=\cw, height=\ch]{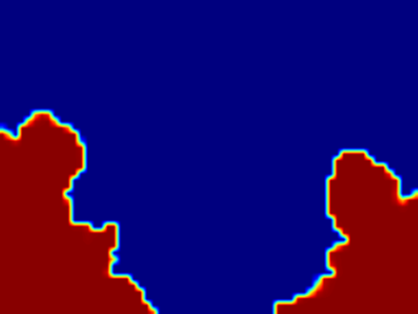} &
        \includegraphics[width=\cw, height=\ch]{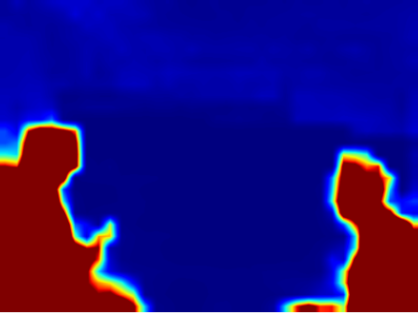} \\
        \multicolumn{6}{c}{\footnotesize TUM \textit{fr3/whs}~\cite{sturm12iros}} \\[4pt]
        \includegraphics[width=\cw, height=\ch]{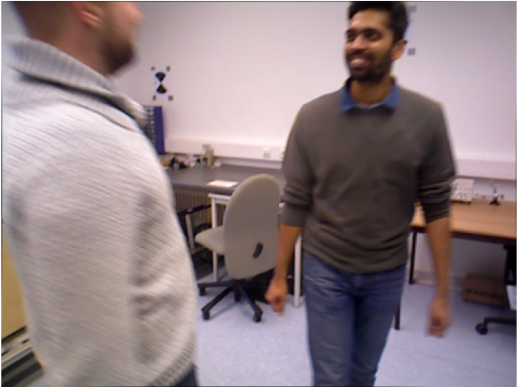} &
        \includegraphics[width=\cw, height=\ch]{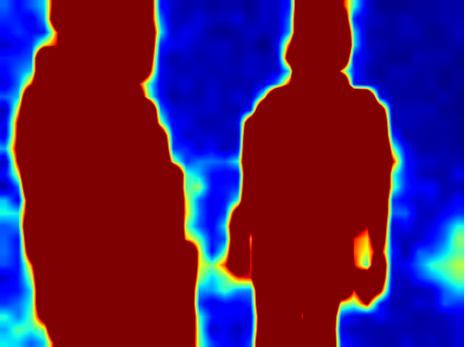} &
        \includegraphics[width=\cw, height=\ch]{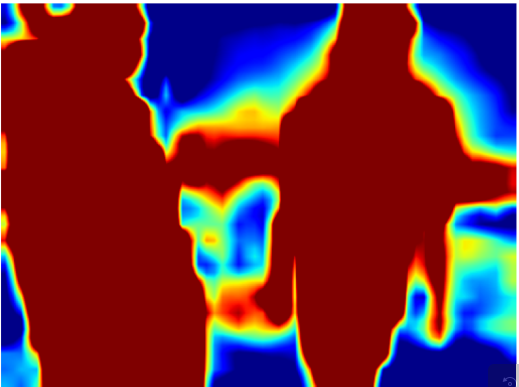} &
        \includegraphics[width=\cw, height=\ch]{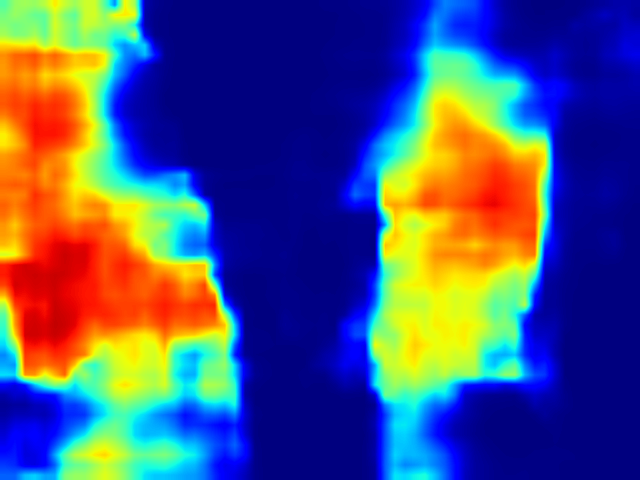} &
        \includegraphics[width=\cw, height=\ch]{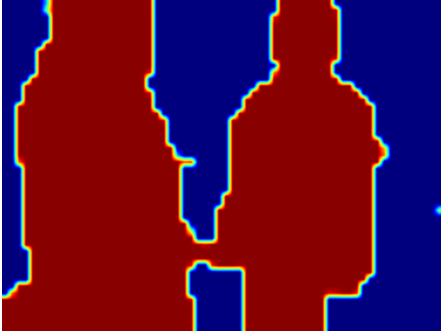} &
        \includegraphics[width=\cw, height=\ch]{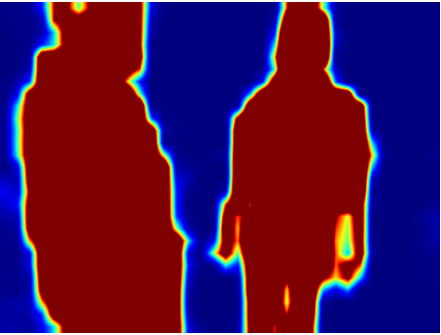} \\
        \multicolumn{6}{c}{\footnotesize Bonn \textit{crowd2}~\cite{refusion3d}} \\[4pt]
        \includegraphics[width=\cw, height=\ch]{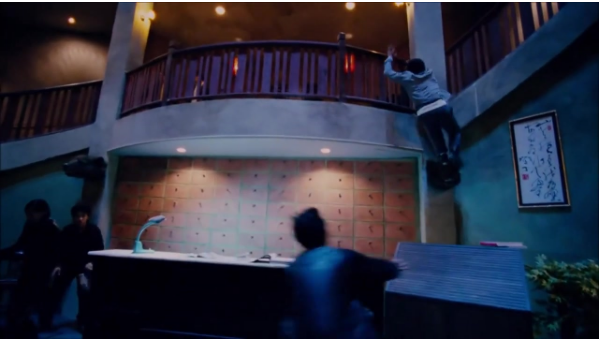} &
        \includegraphics[width=\cw, height=\ch]{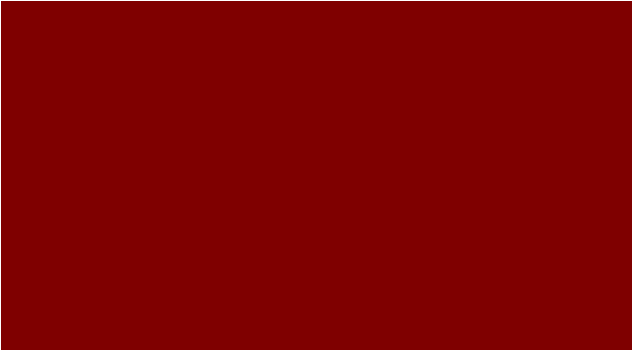} &
        \includegraphics[width=\cw, height=\ch]{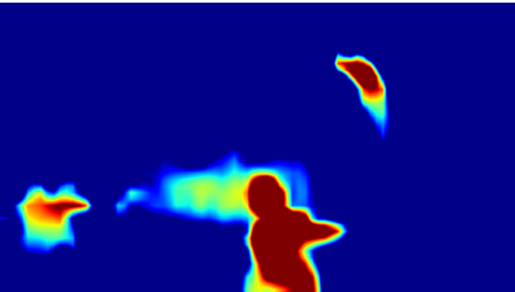} &
        \includegraphics[width=\cw, height=\ch]{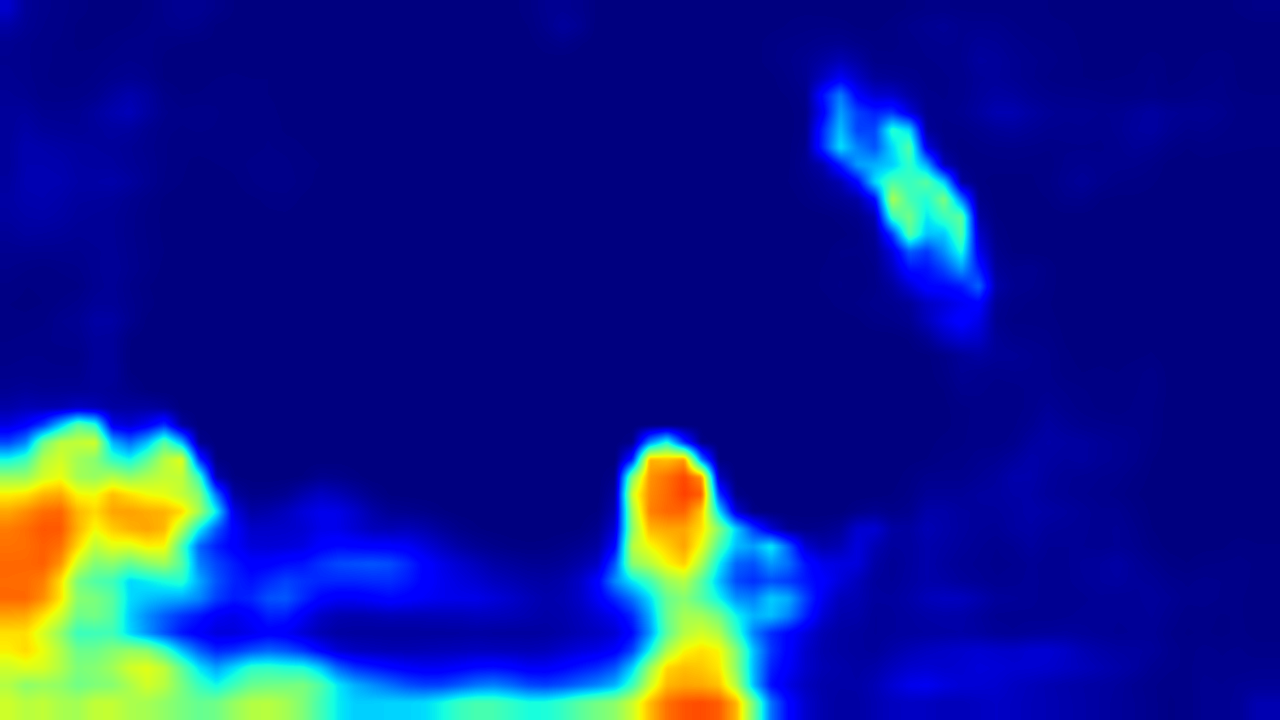} &
        \includegraphics[width=\cw, height=\ch]{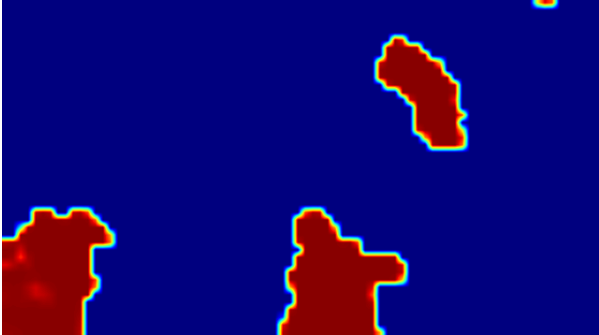} &
        \includegraphics[width=\cw, height=\ch]{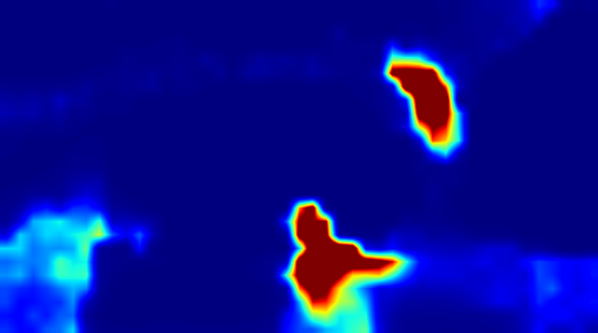} \\
        \multicolumn{6}{c}{\footnotesize Youtube \textit{tomyum}~\cite{li2026droid}} \\
    \end{tabular}
    \vspace{2pt}
    \caption{%
    \textbf{Tracking and mapping uncertainty comparison.}
    Qualitative comparison of uncertainty estimation methods across representative sequences.
    }
    \label{fig:uncer_comparison}
\end{figure*}
\subsection{Tracking Uncertainty Adapter}
\label{subsec:tracking_adapter}
Existing uncertainty-aware 3DGS methods~\cite{zheng2025wildgs, ren2024nerf,zheng2026upslam} train a shallow MLP on rendering residuals, coupling the uncertainty to reconstruction quality. To mitigate this \textit{circular dependency}, the Tracking Uncertainty Adapter (TUA) estimates tracking uncertainty using multi-view feature consistency providing a signal that is not solely dependent on the current reconstruction quality.

\paragraph{Multi-view Feature Consistency (MFC).}
After each BA iteration updates poses $\{\omega_i\}$ and disparities $\{d_i\}$, we measure semantic consistency by warping DINOv2 features $\mathbf{F}_i$ across edges in the frame graph. For each edge $(i, j)$, pixel $p_i$ in frame $i$ is projected into frame $j$ via:
\begin{equation}
\label{eq:warp}
p_{ij} = \Pi_c\!\left(\omega_j^{-1} \omega_i \cdot \Pi_c^{-1}(p_i, d_i)\right)
\end{equation}
The per-pixel MFC error measures feature inconsistency at the warped location:
{\small
\begin{equation}
\label{eq:mfc}
\mathbf{MFC}_{ij}(p) = 1 - \max\big(\cos(\hat{\mathbf{F}}_i(p),\; \operatorname{interp}(\hat{\mathbf{F}}_j, p_{ij})),\; 0\big)
\end{equation}
}%
where $\hat{\mathbf{F}}_i, \hat{\mathbf{F}}_j$ are L2-normalized feature maps and $\operatorname{interp}$ denotes bilinear interpolation. Pixels failing geometric validity checks (negative target depth or out-of-bounds projection) are zeroed out. Moving objects violate the static assumption under rigid-body warping, yielding high MFC error regardless of reconstruction complexity.

\begin{figure*}[t]
    \centering
    \setlength{\tabcolsep}{1.5pt}
    \renewcommand{\arraystretch}{0.5}
    %
    \newcommand{\imgw}{0.121\textwidth}
    \newcommand{\imgh}{0.080\textwidth}
    \begin{tabular}{cc|ccc|ccc}
        \multicolumn{2}{c|}{\footnotesize RGB Input} &
          \multicolumn{3}{c|}{\footnotesize SCOUT-SLAM (\textbf{Ours})} &
          \multicolumn{3}{c}{\footnotesize DROID-W~\cite{li2026droid}} \\[2pt]
        {\footnotesize $t-1$} & {\footnotesize $t$} &
          {\footnotesize iter 1} & {\footnotesize iter 8} & {\footnotesize iter 14} &
          {\footnotesize iter 1} & {\footnotesize iter 14} & {\footnotesize iter 28} \\[2pt]
        \includegraphics[width=\imgw,height=\imgh,keepaspectratio=false]{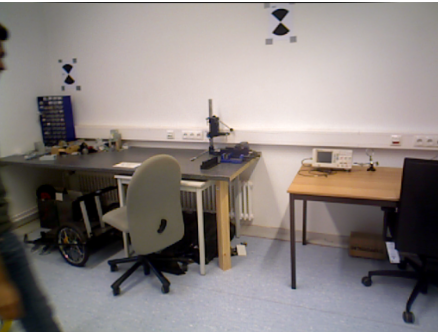} &
        \includegraphics[width=\imgw,height=\imgh,keepaspectratio=false]{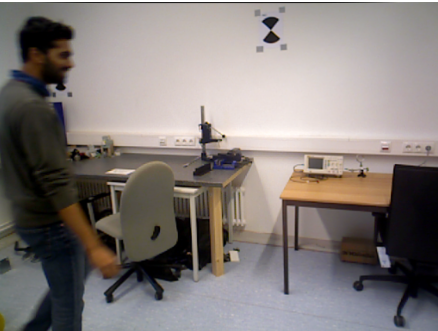} &
        \includegraphics[width=\imgw,height=\imgh,keepaspectratio=false]{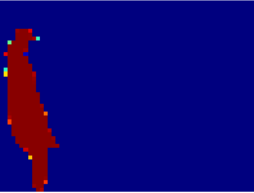} &
        \includegraphics[width=\imgw,height=\imgh,keepaspectratio=false]{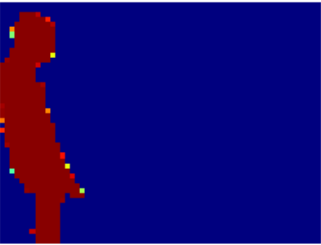} &
        \includegraphics[width=\imgw,height=\imgh,keepaspectratio=false]{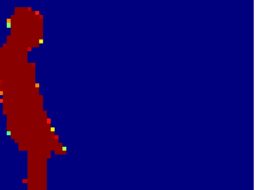} &
        \includegraphics[width=\imgw,height=\imgh,keepaspectratio=false]{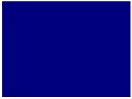} &
        \includegraphics[width=\imgw,height=\imgh,keepaspectratio=false]{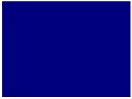} &
        \includegraphics[width=\imgw,height=\imgh,keepaspectratio=false]{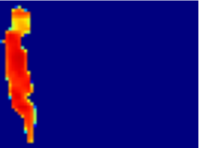} \\
        \multicolumn{2}{c|}{} & \multicolumn{3}{c|}{\makebox[0pt]{\footnotesize Bonn \textit{crowd}~\cite{refusion3d}}} & \multicolumn{3}{c}{} \\[3pt]
        \includegraphics[width=\imgw,height=\imgh,keepaspectratio=false]{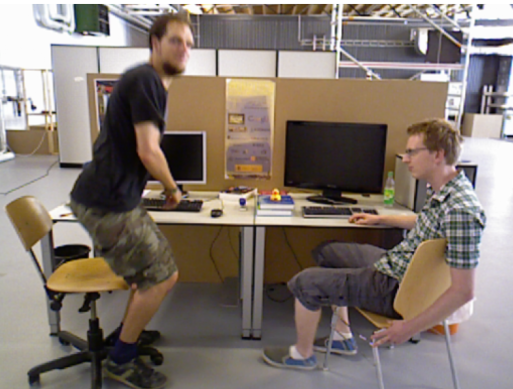} &
        \includegraphics[width=\imgw,height=\imgh,keepaspectratio=false]{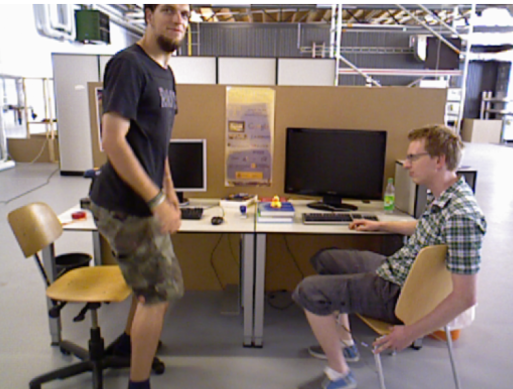} &
        \includegraphics[width=\imgw,height=\imgh,keepaspectratio=false]{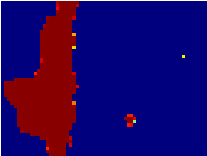} &
        \includegraphics[width=\imgw,height=\imgh,keepaspectratio=false]{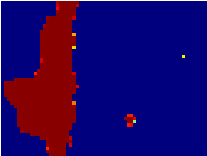} &
        \includegraphics[width=\imgw,height=\imgh,keepaspectratio=false]{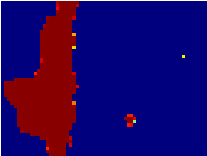} &
        \includegraphics[width=\imgw,height=\imgh,keepaspectratio=false]{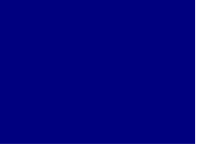} &
        \includegraphics[width=\imgw,height=\imgh,keepaspectratio=false]{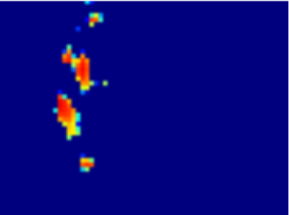} &
        \includegraphics[width=\imgw,height=\imgh,keepaspectratio=false]{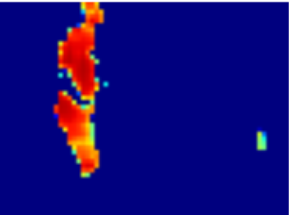} \\
        \multicolumn{2}{c|}{} & \multicolumn{3}{c|}{\makebox[0pt]{\footnotesize TUM \textit{fr3/whs}~\cite{sturm12iros}}} & \multicolumn{3}{c}{} \\[3pt]
    \end{tabular}
    \vspace{2pt}
    \caption{%
    \textbf{Tracking uncertainty convergence across BA iterations.}
    Per-pixel tracking uncertainty maps when dynamic objects enter the field of view. \textbf{Left}: Consecutive input RGB frames. \textbf{Middle}: SCOUT-SLAM isolates moving subjects within early BA iterations (iter 1--14). \textbf{Right}: DROID-W requires more iterations (up to iter 28) to localize dynamic objects. By initializing the Tracking Uncertainty Adapter (TUA) from the shared mapping network, SCOUT-SLAM converges quicker and prevents early-stage dynamic outliers from corrupting pose optimization.
    }
    \label{fig:uncer_convergence}
\end{figure*}

\paragraph{LoRA Architecture.}
To enable bidirectional feedback (\cref{subsec:tracker_conditioned_mapping}), tracking and mapping uncertainty estimation must share a feature space. This is achieved by structuring the TUA as a low-rank adaptation over the MUNet. Given input features $\mathbf{F_i}$ and frozen base weights $\mathbf{W}_0^{(1)}, \mathbf{W}_0^{(2)}$, we inject trainable low-rank matrices $\mathbf{A}^{(1)}, \mathbf{B}^{(1)}$ and $\mathbf{A}^{(2)}, \mathbf{B}^{(2)}$ to compute the tracking uncertainty $\mathbf{u}_{\text{track}_i}$:
\begin{equation}
\label{eq:u_track}
\begin{split}
\mathbf{W}^{(l)} &= \mathbf{W}^{(l)}_0 + \mathbf{B}^{(l)}\mathbf{A}^{(l)}, \quad l \in \{1,2\}\\
\mathbf{u}_{\text{track}_i} &= \text{Softplus}\!\left(\mathbf{W}^{(2)}\,\sigma\!\left(\mathbf{W}^{(1)}\mathbf{F}_i\right)\right)
\end{split}
\end{equation}
where $\sigma(\cdot)$ is $\operatorname{ReLU}$ activation of the MUNet hidden layer. At each BA iteration, the TUA inherits a frozen snapshot of the MUNet, optimizing only low-rank matrices. For the default rank $r{=}8$ used, these matrices introduce only $\sim$5.1k trainable parameters across the entire network. Motion within a local BA window is generally low-dimensional, and optimizing a full MLP on a local BA window risks overfitting to temporary frame alignments. The low-rank bottleneck $r$ constrains updates to primary motion components while preserving the base features of the MUNet. Also, adapting the MUNet, which contains global scene representation, results in faster convergence of tracking uncertainties to mask dynamic entities, as shown in \cref{fig:uncer_convergence}.

\paragraph{Training.}
The LoRA parameters are optimized by minimizing the NLL loss over $\mathbf{MFC}$ observations across all valid edges:
{\small
\begin{equation}
\label{eq:tracking_loss}
\mathcal{L}_\mathbf{MFC} = \frac{1}{\sum\limits_{(i,j) \in \mathcal{E}}|\Omega|}\sum_{(i,j) \in \mathcal{E}} \sum_{p \in \Omega_{ij}} \frac{\mathbf{MFC}_{ij}(p)}{\mathbf{u_{\text{track}_i}}^2(p)}
+ \log \mathbf{u_{\text{track}_i}}(p)
\end{equation}}%
where $\mathcal{E}$ is the edge set in the current BA window, $\Omega_{ij}$ is the set of geometrically valid pixels for edge $(i,j)$, and $\mathbf{u_{\text{track}_i}}(p) > 0$ is enforced by softplus. $\log \mathbf{u_{\text{track}_i}}$ prevents $\mathbf{u_{\text{track}_i}} \to \infty$. The loss is optimized in a single gradient step per BA iteration.

\subsection{Uncertainty-Guided Dense Bundle Adjustment}
The TUA's uncertainty is integrated into a dense BA framework. To ensure numerical stability and provide a lower bound for dynamic pixels, the tracking uncertainty $\mathbf{u_{\text{track}_i}}$ is converted to a per-pixel confidence weight $w$:
{\small
\begin{equation}
\label{eq:weight_conversion_w}
w_i = \operatorname{clamp}\!\left(\frac{1}{{\mathbf{u_{\text{track}_i}}}^2},\; \epsilon,\; 1.0\right)
\end{equation}
}%
The confidence $w_i$ modulated the covariance $\Sigma_{ij}$ of the optical flow $\tilde{f}_{ij}$~\cite{teed2021droid} in the ConvGRU. Dividing by $w_i$ inflates the covariance at low-confidence pixels, reducing their influence on the reprojection objective in \cref{eq:dba}. The resulting confidence map is subsequently propagated to the mapping thread.
Camera poses $\omega$ and disparities $d$ are optimized by minimizing the uncertainty-weighted reprojection error. The reprojection error is regularized by a scale-aligned monocular metric depth prior $\tilde{D}_i$~\cite{hu2024metric3dv2} following \cite{zheng2025wildgs,li2026droid}, where $M_i$ is a multi-view depth consistency mask:
{\small
\begin{equation}
\label{eq:dba}
\begin{split}
\underset{\omega, d}{\arg\min} \sum_{(i,j) \in \mathcal{E}} &\left\| \tilde{f}_{ij} - \Pi_c\!\left(\omega_j^{-1} \omega_i\, \Pi_c^{-1}(p_i, d_i)\right) \right\|^2_{\Sigma_{ij} / w_i} \\
&+ \mu_m \sum_{i \in \mathcal{V}} \left\| M_i (d_i - 1/\tilde{D}_i) \right\|^2. 
\end{split}
\end{equation}
}%
\\
To avoid memory bottlenecks when warping 384-dim DINOv2 features for the MFC loss (\cref{eq:mfc}) and depth mask $M_i$, custom CUDA kernels are implemented that bypass intermediate tensor allocations.

\subsection{Spatially-Adaptive Prior}
\label{subsec:tracker_conditioned_mapping}
The MUNet takes the pretrained DINOv2 features $\mathbf{F_i}$ as input to predict the per-pixel mapping uncertainty $\mathbf{u_{\text{map}_i}}$. \cref{eq:u_map} and \cref{eq:mlp_loss} denote element-wise operations over the spatial dimensions. Operating strictly on the base layers $W_0$, the mapping uncertainty $\mathbf{u_{\text{map}_i}}$ is computed as:
{\small
\begin{equation}
\label{eq:u_map}
\mathbf{u}_{\text{map}_i} = \text{Softplus}\!\left(\mathbf{W}^{(2)}_0\,\sigma\!\left(\mathbf{W}^{(1)}_0\mathbf{F}_i\right)\right)
\end{equation}
}%
Prior works~\cite{zheng2025wildgs, ren2024nerf} train MUNet using a negative log-likelihood (NLL) with a uniform $\log \mathbf{u}_{\text{map}_i}$ penalty, which blindly inflates uncertainty when residuals are high on challenging static regions. To address this, a spatially-adaptive prior replaces the uniform penalty with $\lambda(w,R)$ conditioned on the TUA's tracking confidence $w$:
{\small
\begin{equation}
\label{eq:adaptive_prior}
\mathcal{L}_\text{uncer} = \frac{\mathcal{L}'_\text{SSIM} + \mu_d \mathcal{L}_\text{depth}}{\mathbf{u^2_{\text{map}_i}}} + \underbrace{\bigl(\lambda_0 + \gamma w \cdot g(R)\bigr)}_{\lambda(w,R)} \cdot \log \mathbf{u_{\text{map}_i}}
\end{equation}
}%
where $\mathcal{L}'_\text{SSIM}$ is the modified SSIM~\cite{zheng2025wildgs} and $\mathcal{L}_\text{depth} = |\mathbf{\hat{D}} - \mathbf{\tilde{D}}|_1$. Since $w \in [0.01, 1.0]$, the adaptive prior $\lambda(w, R)$ is bounded within $[\lambda_0, \lambda_0 + \gamma]$. Primarily, $\lambda$ regularizes uncertainty inflation by forcing $\mathbf{u}_{\text{map}_i}$ to remain low when confidence $w$ is high. To prevent the map from overfitting to transient variations (e.g., sensor noise or lighting flicker) in these static regions, the combined rendering residual $R = \mathcal{L}'_\text{SSIM} + \mu_d \mathcal{L}_\text{depth}$ is not too large, it modulates the prior via a gating mechanism $g(R) = \exp(-\alpha R)$, preventing the 3DGS map from being forced to render sensor noise. Refer to supplementary for more details. Additionally, the spatial-adaptive prior shields the shared base weights $\mathbf{W}_0$ from noisy gradients from $\mathcal{L_\text{uncer}}$, maintaining a stable feature space for TUA. We also apply a variance regularization $\mathcal{L}_\text{reg}$~\cite{ren2024nerf} to encourage smooth uncertainty across similar DINOv2 features. With hyperparameters $\mu_{*}$, the total loss is:
{\small
\begin{equation}
\label{eq:mlp_loss}
\mathcal{L}_\text{MUNet} = \mathcal{L}_\text{uncer} + \mu_\text{reg} \mathcal{L}_\text{reg}
\end{equation}
}%

\subsection{3DGS Map Optimization}
\label{subsec:map_optimization}
At each mapping iteration, a keyframe is sampled with equal probability from either the spatial co-visibility window or the historical keyframe buffer. \cref{eq:render_loss} and \cref{eq:color_loss} denote element-wise operations over the spatial dimensions. The scene is then rendered and the Gaussian attributes are optimized by minimizing:
{\small
\begin{equation}
\label{eq:render_loss}
\mathcal{L}_\text{render} = \frac{\mu_\text{rgb} \mathcal{L}_\text{color} + (1-\mu_\text{rgb} ) \mathcal{L}_\text{depth}}{\mathbf{u_{\text{map}_i}}^2} + \mu_\text{iso} \mathcal{L}_\text{iso}
\end{equation}
}%
The color loss combines $L_1$ and SSIM terms following~\cite{kerbl20233d,zheng2025wildgs}:
{\small
\begin{equation}
    \mathcal{L}_\text{color} = (1 - \mu_\text{SSIM}) \|\hat{C} - C\|_1 + \mu_\text{SSIM} \mathcal{L}_\text{SSIM}
    \label{eq:color_loss}
\end{equation}
}%
The depth loss $\mathcal{L}_\text{depth}$ penalizes deviation from the monocular metric depth prior~\cite{hu2024metric3dv2}. 
Dividing the residuals by the detached mapping uncertainty $\mathbf{u_{\text{map}_i}}^2$ ensures $\mathcal{L_\text{render}}$ only optimizes the gaussians, esnsuring MUNet and 3D gaussians are optimized in parallel.  Consequently, the gaussians are optimized to only reconstruct the static scene. Finally, the isotropic regularization $\mathcal{L}_\text{iso}$~\cite{Matsuki_2024_CVPR} constrains Gaussian scales to prevent excessive elongation in sparsely observed regions. 
\begin{figure*}[t]
    \centering
    \setlength{\tabcolsep}{1.5pt}
    \renewcommand{\arraystretch}{0.5}
    \newcommand{\imgw}{0.19\textwidth}
    \newcommand{\imghOne}{0.095\textwidth}
    \newcommand{\imghTwo}{0.095\textwidth}
    \newcommand{\imghThree}{0.095\textwidth}
    \begin{tabular}{cc|c|cc}
        {\scriptsize \shortstack{Mapping Uncer.\\(WildGS-SLAM~\cite{zheng2025wildgs})}} & 
        {\scriptsize \shortstack{Mapping Uncer.\\(\textbf{Ours})}} & 
        {\scriptsize \shortstack{Input\\Image}} & 
        {\scriptsize \shortstack{Rendered static image\\(\textbf{Ours})}} & 
        {\scriptsize \shortstack{Rendered\\(WildGS-SLAM~\cite{zheng2025wildgs})}} \\[3pt]
        \includegraphics[width=\imgw,height=\imghOne,keepaspectratio=false]{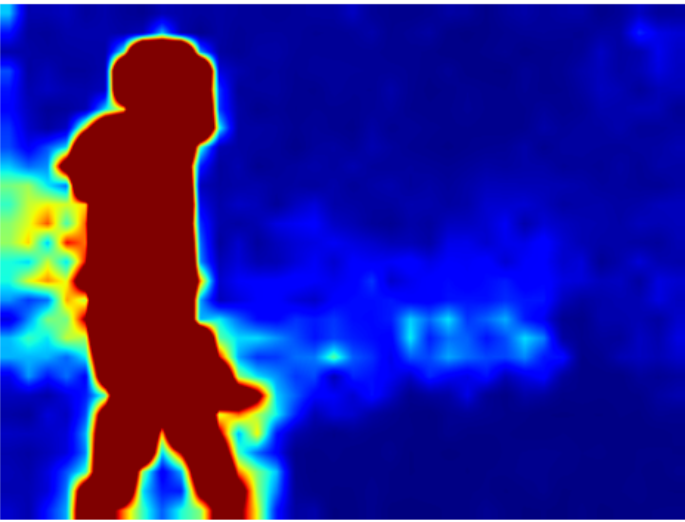} &
        \includegraphics[width=\imgw,height=\imghOne,keepaspectratio=false]{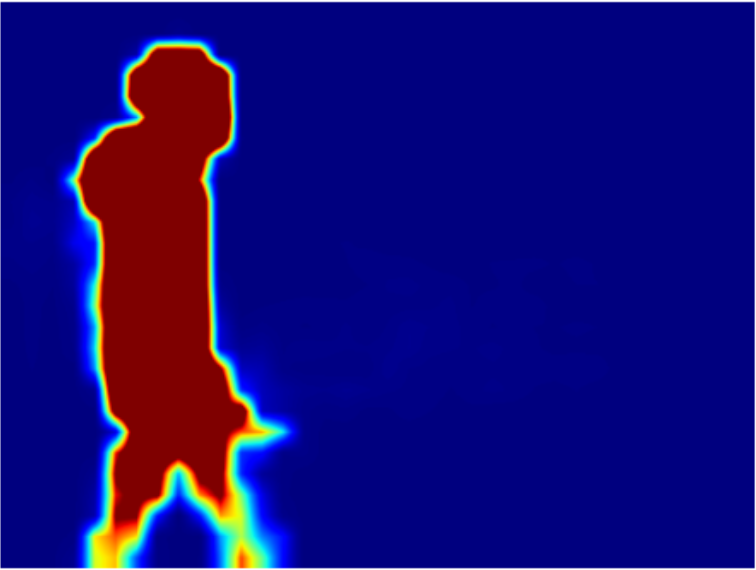} &
        \includegraphics[width=\imgw,height=\imghOne,keepaspectratio=false]{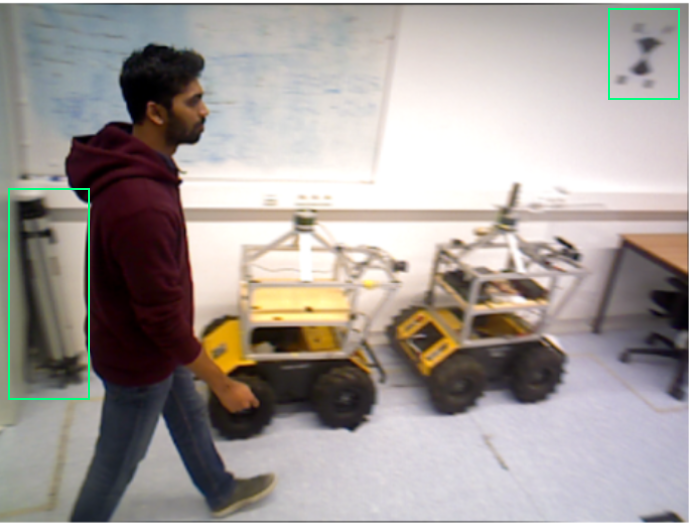} &
        \includegraphics[width=\imgw,height=\imghOne,keepaspectratio=false]{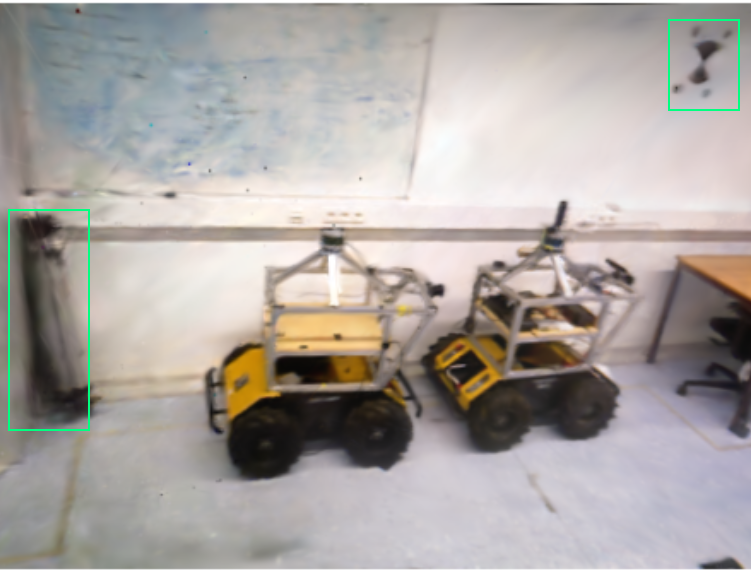} &
        \includegraphics[width=\imgw,height=\imghOne,keepaspectratio=false]{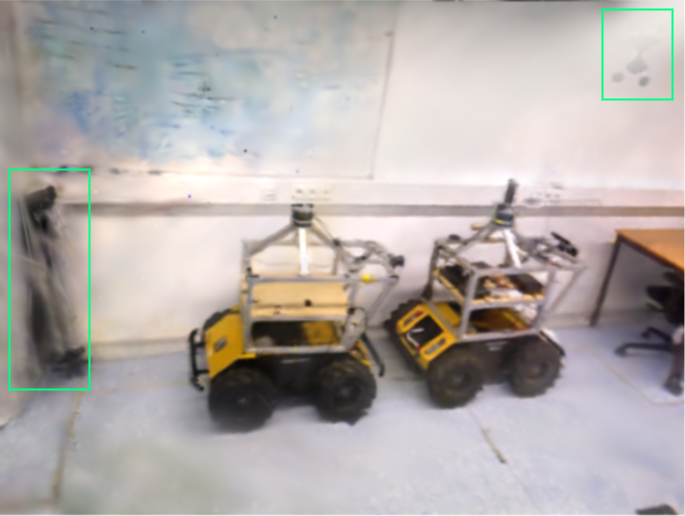} \\
        \multicolumn{2}{c|}{} & \multicolumn{1}{c|}{\makebox[0pt]{\footnotesize Bonn \textit{person\_tracking2}~\cite{refusion3d}}} & \multicolumn{2}{c}{} \\[4pt]
        \includegraphics[width=\imgw,height=\imghTwo,keepaspectratio=false]{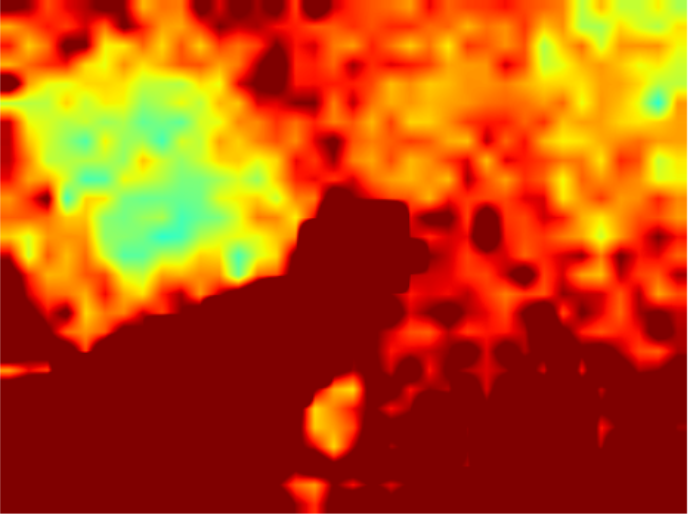} &
        \includegraphics[width=\imgw,height=\imghTwo,keepaspectratio=false]{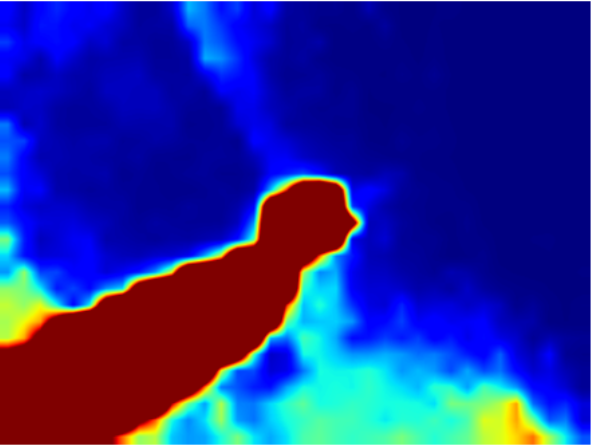} &
        \includegraphics[width=\imgw,height=\imghTwo,keepaspectratio=false]{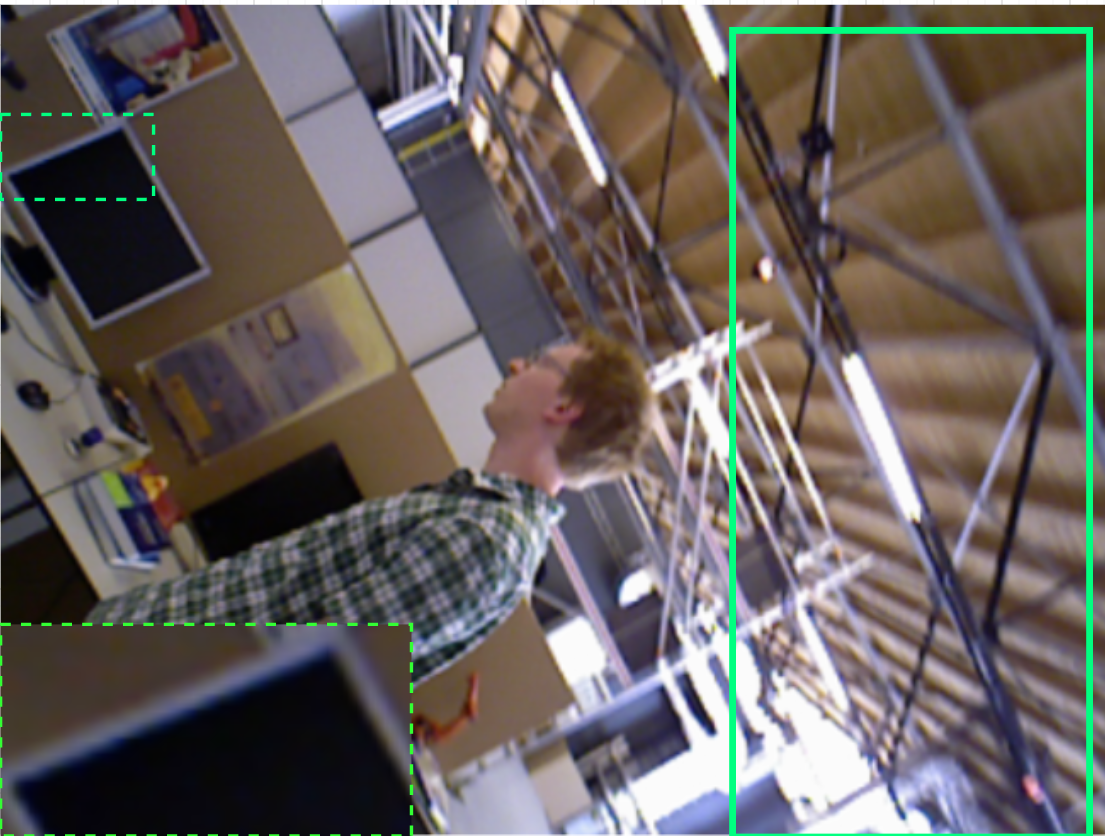} &
        \includegraphics[width=\imgw,height=\imghTwo,keepaspectratio=false]{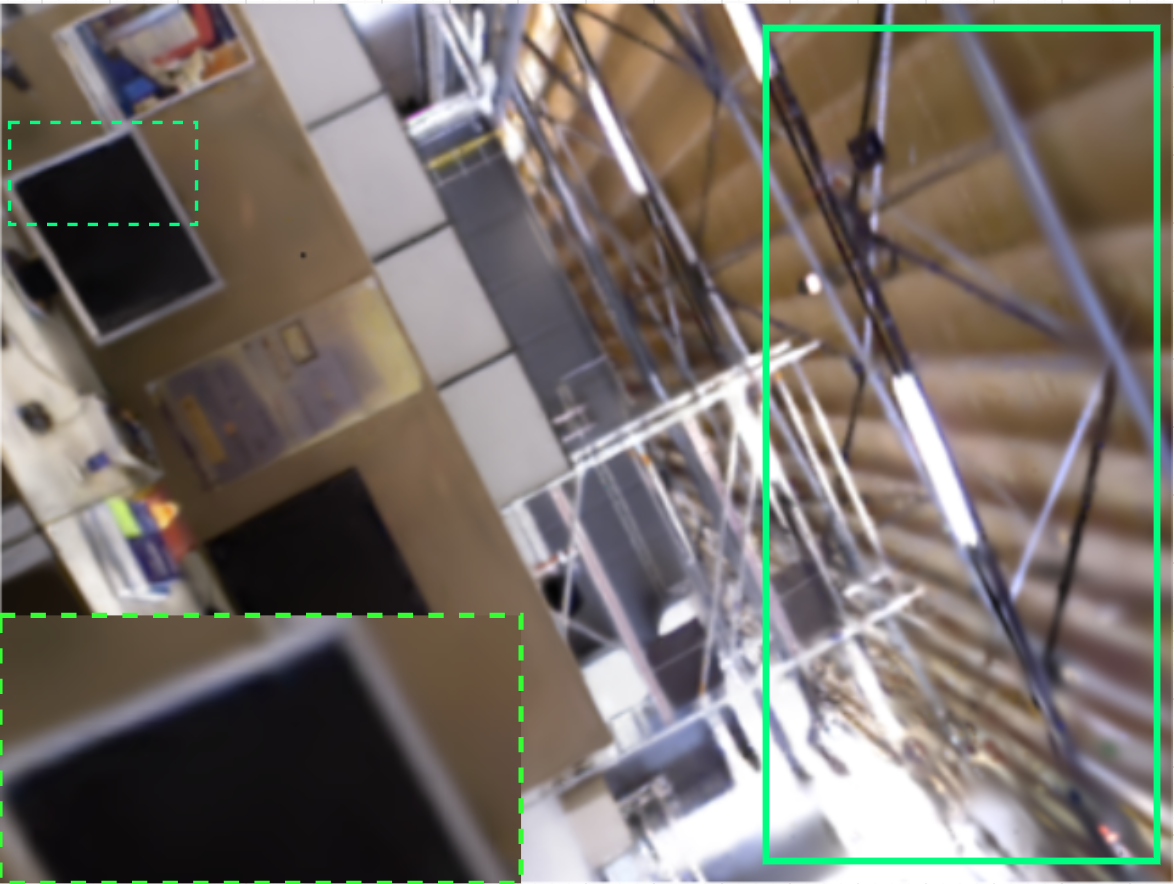} &
        \includegraphics[width=\imgw,height=\imghTwo,keepaspectratio=false]{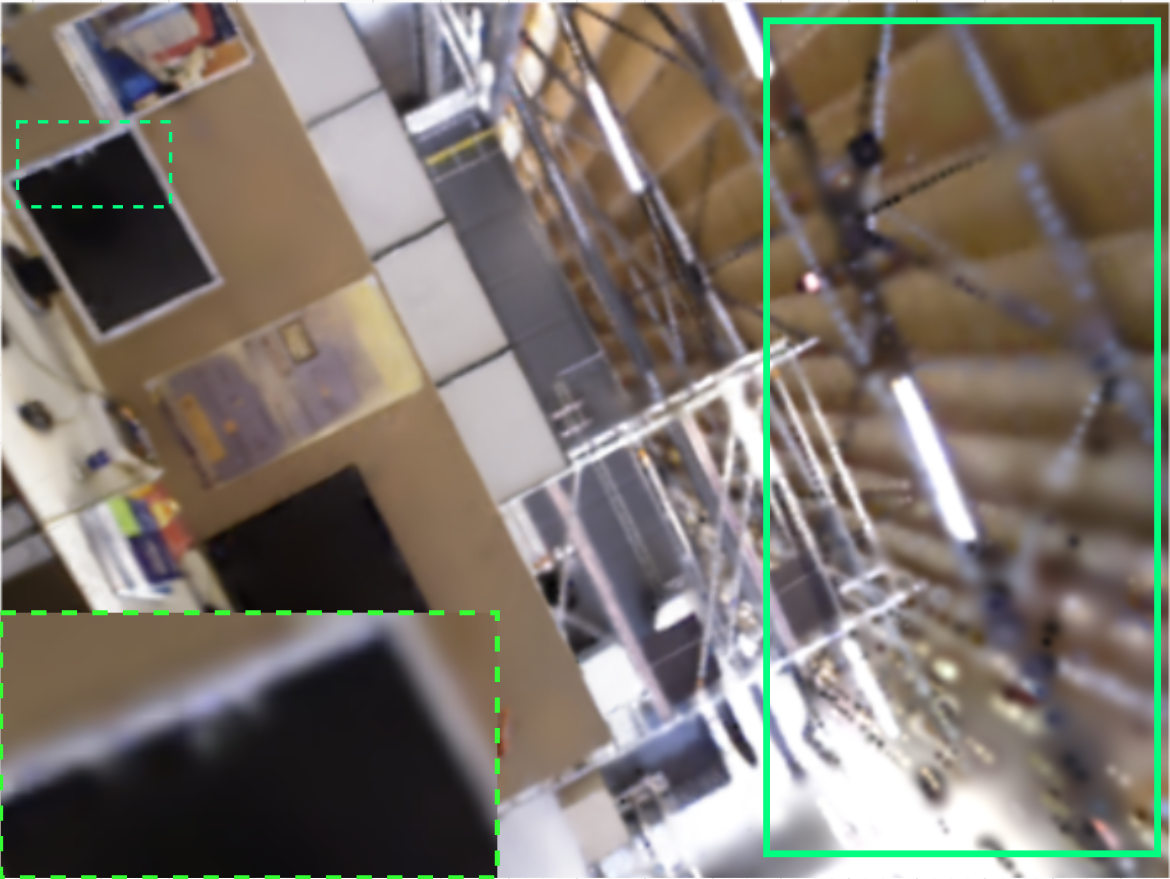} \\
        \multicolumn{2}{c|}{} & \multicolumn{1}{c|}{\makebox[0pt]{\footnotesize TUM \textit{fr3/wr}~\cite{sturm12iros}}} & \multicolumn{2}{c}{} \\[4pt]
        \includegraphics[width=\imgw,height=\imghThree,keepaspectratio=false]{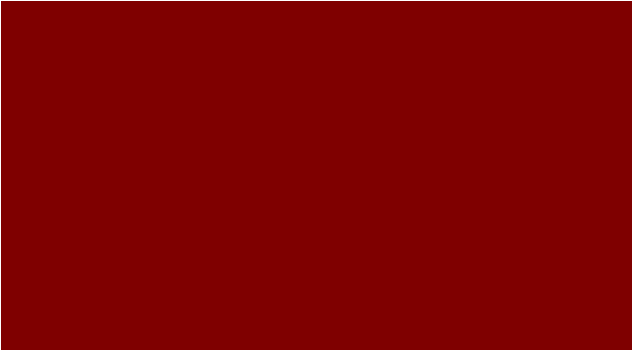} &
        \includegraphics[width=\imgw,height=\imghThree,keepaspectratio=false]{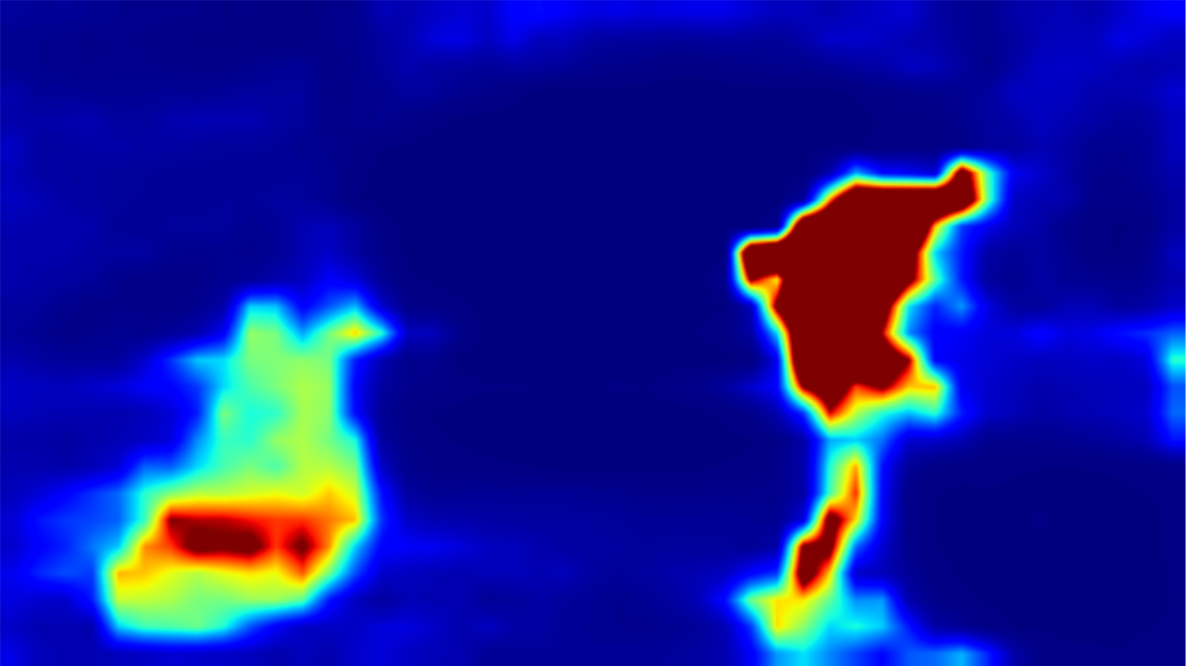} &
        \includegraphics[width=\imgw,height=\imghThree,keepaspectratio=false]{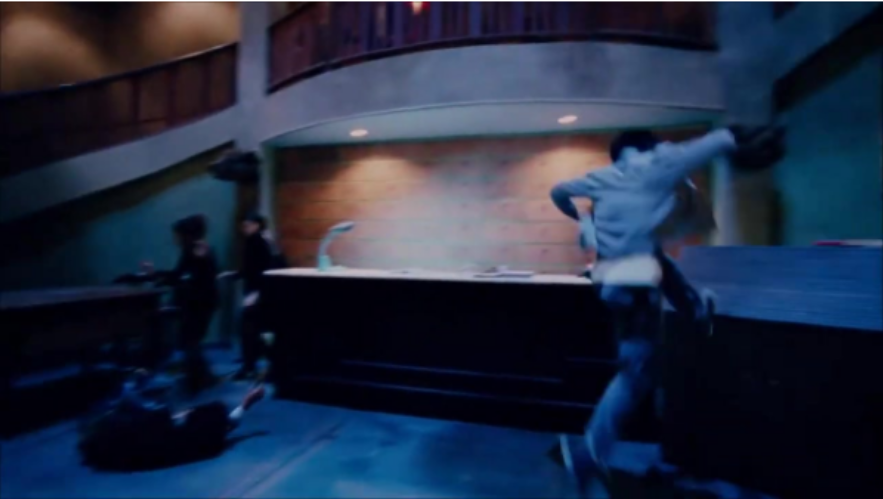} &
        \includegraphics[width=\imgw,height=\imghThree,keepaspectratio=false]{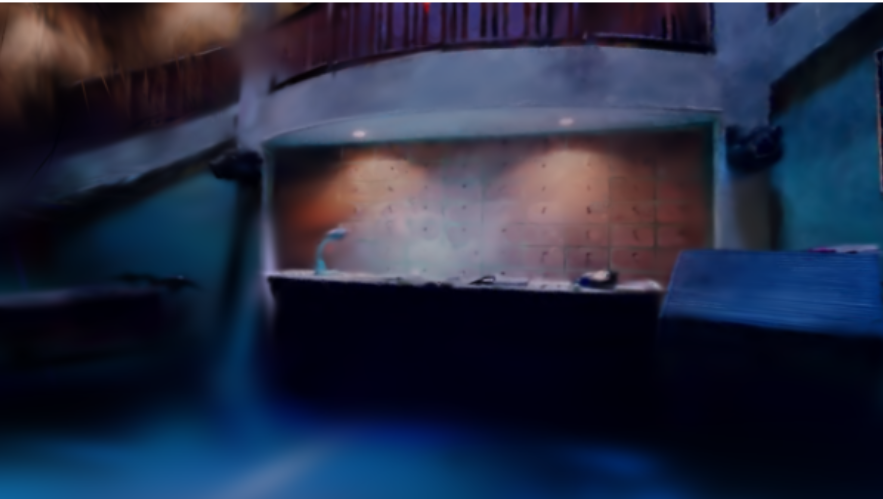} &
        \includegraphics[width=\imgw,height=\imghThree,keepaspectratio=false]{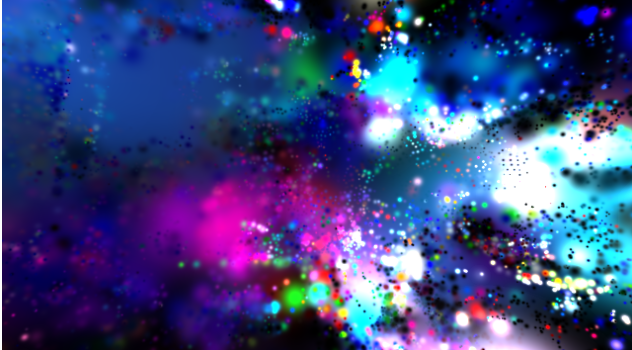} \\
        \multicolumn{2}{c|}{} & \multicolumn{1}{c|}{\makebox[0pt]{\footnotesize Youtube \textit{tomyum}~\cite{li2026droid}}} & \multicolumn{2}{c}{} \\
    \end{tabular}
    \vspace{2pt}
    \caption{%
    \textbf{Comparison of Rendered static image and mapping uncertainty.} \textbf{Left}: Mapping uncertainties. \textbf{Middle}: Input RGB. \textbf{Right}: Rendered static images. WildGS-SLAM assigns erroneously high uncertainty to static regions, whereas SCOUT-SLAM's mapping uncertainties accurately isolate dynamic objects, resulting in crisper, artifact-free static reconstructions. The solid and dashed green boxes highlight the preserved thin structures and reduced smearing in SCOUT-SLAM's reconstruction. \textbf{Top row}: Challenging lighting changes in Bonn \textit{person\_tracking2}. \textbf{Middle row}: Complex static structural geometry in TUM \textit{fr3/wr}. \textbf{Bottom row}: Fast camera motion in Youtube \textit{tomyum} degrades the WildGS-SLAM's reconstruction, whereas the spatial-adaptive prior ensures reconstruction stability in SCOUT-SLAM.
    }
    \label{fig:photo_cir_dep}
\end{figure*}

\begin{table*}[t]
\centering
\footnotesize
\setlength{\tabcolsep}{4.5pt}
{
\begin{tabular}{lccccccccc}
\toprule
Method & \texttt{balloon} & \texttt{balloon2} & \texttt{crowd} & \texttt{crowd2} & \texttt{mov\_box} & \texttt{mov\_box2} & \texttt{person} & \texttt{person2} & Avg. \\
\midrule
\multicolumn{10}{l}{\cellcolor[HTML]{EEEEEE}{\textit{SLAM}}} \\
DROID-W \cite{li2026droid} & \cellcolor{thirdc}2.7 & \cellcolor{thirdc}2.5 & \cellcolor{firstc}\textbf{1.3} & \cellcolor{secondc}1.8 & \cellcolor{thirdc}1.6 & \cellcolor{firstc}\textbf{2.3} & \cellcolor{thirdc}3.5 & \cellcolor{secondc}3.0 & \cellcolor{secondc}2.32 \\
DROID-SLAM \cite{teed2021droid} & 7.5 & 4.1 & 5.2 & 6.5 & 2.3 & 4.0 & 4.3 & 5.4 & 4.91 \\
MegaSaM \cite{li2025megasam} & 3.7 & 2.6 & \cellcolor{thirdc}1.6 & 7.2 & \cellcolor{secondc}1.4 & 3.4 & 4.1 & 4.0 & 3.51 \\
\hdashline
\noalign{\vskip 1pt}
\multicolumn{10}{l}{\cellcolor[HTML]{EEEEEE}{\textit{Neural \& 3DGS SLAM}}} \\
DynaMoN \cite{schischka2024dynamon} & 2.8 & 2.7 & 3.5 & 2.8 & \cellcolor{firstc}\textbf{1.3} & 2.7 & 14.8 & \cellcolor{firstc}\textbf{2.2} & 4.10 \\
DDN-SLAM \cite{li2025ddn} & \cellcolor{firstc}\textbf{1.8} & 4.1 & 1.8 & 2.3 & 2.0 & 3.2 & 4.3 & 3.8 & 2.91 \\
RoDyn-SLAM \cite{jiang2024rodyn} & 7.9 & 11.5 & - & - & - & 12.6 & 14.5 & 13.8 & -\\
Splat-SLAM \cite{Sandstrom_2025_CVPR} & 8.8 & 3.0 & 6.8 & F & 1.7 & 3.0 & 4.9 & 25.8 & F \\
UP-SLAM~\cite{zheng2026upslam} & 2.8 & 2.7 & - & - & - & 3.2 & 4.0 & 3.6 & -\\
WildGS-SLAM \cite{zheng2025wildgs} & \cellcolor{thirdc}2.7 & \cellcolor{secondc}2.4 & \cellcolor{thirdc}1.6 & \cellcolor{thirdc}2.2 & 1.7 & \cellcolor{thirdc}2.5 & \cellcolor{thirdc}3.5 & \cellcolor{thirdc}3.1 & \cellcolor{thirdc}2.47 \\
\hdashline
SCOUT-SLAM (\textbf{Ours}) & \cellcolor{secondc}2.4 & \cellcolor{firstc}\textbf{2.2} & \cellcolor{secondc}1.4 & \cellcolor{firstc}\textbf{1.7} & \cellcolor{thirdc}1.6 & \cellcolor{secondc}2.4 & \cellcolor{secondc}3.4 & \cellcolor{thirdc}3.1 & \cellcolor{firstc}\textbf{2.29} \\
\bottomrule
\end{tabular}
}
\vspace{-2mm}
\caption{\textbf{Tracking Results on Bonn Dynamic}~\cite{refusion3d}. ATE RMSE [cm]. Best results are highlighted as \colorbox{firstc}{\textbf{first}}, \colorbox{secondc}{second}, and \colorbox{thirdc}{third}. 'F' indicated tracking failure. '-' denotes unreported sequences.}
\label{tab:bonn_results}
\vspace{-5pt}
\end{table*}

\section{Experiments}
\label{sec:experiments}
\subsection{Experimental Setup}
\noindent\textbf{Datasets.}
Evaluation is conducted across three dynamic benchmarks: TUM RGB-D~\cite{sturm12iros}, Bonn Dynamic~\cite{refusion3d}, Wild-SLAM MoCap~\cite{zheng2025wildgs}, qualitative evaluations on dynamic videos from \textit{Youtube}~\cite{li2026droid}.  Refer to supplementary for further details.
\\
\noindent\textbf{Baselines.}
We evaluate SCOUT-SLAM against SOTA dynamic SLAM methods, categorized into two paradigms: (1) \textbf{Standard SLAM} methods: including DROID-SLAM~\cite{teed2021droid}, MegaSaM~\cite{li2025megasam}, DROID-W~\cite{li2026droid}. (2) \textbf{Neural \& 3DGS SLAM} methods: including NICE-SLAM~\cite{zhu2022nice},  DynaMoN~\cite{schischka2024dynamon}, DDN-SLAM~\cite{li2025ddn}, RoDyn-SLAM~\cite{jiang2024rodyn}, Splat-SLAM~\cite{Sandstrom_2025_CVPR}, UP-SLAM~\cite{zheng2026upslam}, and WildGS-SLAM~\cite{zheng2025wildgs}.
\\
\noindent\textbf{Implementation \& Metrics}
SCOUT-SLAM is evaluated on an Intel Core i9 CPU and RTX 4090 GPU, utilizing standard ATE RMSE for tracking, alongside PSNR, SSIM, and LPIPS for view synthesis. Dynamic objects are masked out using SAM3~\cite{carion2025sam3segmentconcepts}. For comprehensive details regarding the SLAM pipeline, network architectures, hyperparameters, and extended evaluation masking, please refer to the supplementary material.
\begin{table}[!t]
\centering
\footnotesize
\setlength{\tabcolsep}{3.5pt}
\resizebox{\columnwidth}{!}{
\begin{tabular}{lcccccc}
\toprule
Method & \texttt{fr2/dp} & \texttt{fr3/whs} & \texttt{fr3/wr} & \texttt{fr3/ws} & \texttt{fr3/wx} & Avg. \\
\midrule
\multicolumn{7}{l}{\cellcolor[HTML]{EEEEEE}{\textit{SLAM}}} \\
\scriptsize DROID-W \cite{li2026droid} & 1.2 & \cellcolor{thirdc}1.6 & \cellcolor{thirdc}3.1 & \cellcolor{secondc}0.5 & \cellcolor{firstc}\textbf{1.2} & \cellcolor{thirdc}1.52 \\
\scriptsize DROID-SLAM \cite{teed2021droid} & \cellcolor{firstc}\textbf{0.6} & 2.2 & 4.0 & 1.2 & 1.6 & 1.92 \\
\scriptsize MegaSaM \cite{li2025megasam} & 1.6 & 1.8 & \cellcolor{firstc}\textbf{2.6} & \cellcolor{thirdc}0.6 & 1.5 & 1.62 \\
\hdashline
\noalign{\vskip 1pt}
\multicolumn{7}{l}{\cellcolor[HTML]{EEEEEE}{\textit{Neural \& 3DGS SLAM}}} \\
\scriptsize DynaMoN\cite{schischka2024dynamon} & \cellcolor{secondc}0.7 & 1.9 & 3.9 & 0.7 & \cellcolor{thirdc}1.4 & 1.72 \\
\scriptsize DDN-SLAM \cite{li2025ddn} & - & 2.3 & 3.9 & 1.0 & \cellcolor{secondc}1.3 & - \\
\scriptsize RoDyn-SLAM \cite{jiang2024rodyn} & - & 5.6 & - & 1.7 & 8.3 & - \\
\scriptsize Splat-SLAM~\cite{Sandstrom_2025_CVPR} & \cellcolor{secondc}0.7 & 2.2 & 3.9 & 2.3 & \cellcolor{secondc}1.3 & 2.08 \\
\scriptsize UP-SLAM~\cite{zheng2026upslam} & 1.3 & 2.6 & - & 0.7 & 1.6 & - \\
\scriptsize WildGS-SLAM \cite{zheng2025wildgs} & 1.3 & \cellcolor{firstc}\textbf{1.4} & \cellcolor{thirdc}3.1 & \cellcolor{secondc}0.5 & \cellcolor{firstc}\textbf{1.2} & \cellcolor{secondc}1.51 \\
\hdashline
\scriptsize SCOUT-SLAM(\textbf{Ours}) & \cellcolor{thirdc}1.1 & \cellcolor{secondc}1.5 & \cellcolor{secondc}2.9 & \cellcolor{firstc}\textbf{0.4} & \cellcolor{firstc}\textbf{1.2} & \cellcolor{firstc}\textbf{1.42} \\
\bottomrule
\end{tabular}
}
\vspace{-2mm}
\caption{\textbf{Tracking Results on TUM RGB-D\cite{sturm12iros} Dynamic Sequences.} ATE RMSE [cm]. Best results are highlighted as \colorbox{firstc}{\textbf{first}}, \colorbox{secondc}{second}, and \colorbox{thirdc}{third}. '-' denotes unreported sequences.}
\label{tab:tum_results}
\vspace{-2em}
\end{table}


\begin{table*}[t]
\centering
\footnotesize
\renewcommand{\arraystretch}{0.85}
\setlength{\tabcolsep}{4.5pt}
{
\begin{tabular}{llcccccccccc}
\toprule
Method & Metric & \texttt{fr2/dp} & \texttt{fr3/shs} & \texttt{fr3/sr} & \texttt{fr3/ss} & \texttt{fr3/sx} & \texttt{fr3/whs} & \texttt{fr3/wr} & \texttt{fr3/ws} & \texttt{fr3/wx} & Avg. \\
\midrule
\scriptsize Splat-SLAM \cite{Sandstrom_2025_CVPR} & PSNR $\uparrow$ & 17.87 & 17.14 & 16.65 & \textbf{24.20} & 19.81 & 14.68 & 13.84 & 15.24 & \textbf{16.34} & 17.31 \\
 & SSIM $\uparrow$ & 0.632 & 0.625 & 0.678 & 0.862 & 0.727 & 0.513 & 0.480 & 0.610 & 0.597 & 0.636 \\
 & LPIPS $\downarrow$ & 0.469 & 0.423  & 0.336 & 0.147 & 0.235 & 0.497 & 0.641 & 0.460 & 0.422 & 0.404 \\
\hdashline
\scriptsize WildGS-SLAM \cite{zheng2025wildgs} & PSNR $\uparrow$ & 21.36 & 18.62 & 20.59 & 23.62 & 21.53 & 16.06 & 15.44 & 17.55 & 15.46 & 18.91 \\
 & SSIM $\uparrow$ & 0.749 & 0.724 & 0.769 & 0.869 & 0.792 & 0.660 & 0.661 & 0.774 & 0.657 & 0.739 \\
 & LPIPS $\downarrow$ & 0.201 & 0.264 & 0.208 & 0.108 & 0.172 & 0.337 & 0.350 & 0.205 & 0.304 & 0.239 \\
\hdashline
\noalign{\vskip 1pt}
\scriptsize SCOUT-SLAM (\textbf{Ours}) & PSNR $\uparrow$ & \textbf{21.90} & \textbf{19.28} & \textbf{20.88} & 24.01 & \textbf{22.03} & \textbf{16.59} & \textbf{15.63} & \textbf{17.72} & 16.03 & \textbf{19.34} \\
 & SSIM $\uparrow$ & \textbf{0.782} & \textbf{0.742} & \textbf{0.787} & \textbf{0.877} & \textbf{0.808} & \textbf{0.694} & \textbf{0.674} & \textbf{0.784} & \textbf{0.703} & \textbf{0.761} \\
 & LPIPS $\downarrow$ & \textbf{0.175} & \textbf{0.243} & \textbf{0.190} & \textbf{0.103} & \textbf{0.165} & \textbf{0.293} & \textbf{0.335} & \textbf{0.201} & \textbf{0.267} & \textbf{0.219} \\
\bottomrule
\end{tabular}
}
\vspace{-3mm}
\caption{\textbf{Input View Synthesis on TUM RGB-D.} PSNR$\uparrow$ / SSIM$\uparrow$ / LPIPS$\downarrow$. Best results are in \textbf{bold}}
\label{tab:tum_mapping}
\vspace{-6pt}
\end{table*}

\begin{table*}[t]
\centering
\footnotesize
\setlength{\tabcolsep}{4.5pt}
{
\begin{tabular}{lccccccccccc}
\toprule
Method & \texttt{ANYmal1} & \texttt{ANYmal2} & \texttt{ball} & \texttt{crowd} & \texttt{person} & \texttt{racket} & \texttt{stones} & \texttt{table1} & \texttt{table2} & \texttt{umbrella} & Avg. \\
\midrule
\multicolumn{12}{l}{\cellcolor[HTML]{EEEEEE}{\textit{SLAM}}} \\
DROID-W \cite{li2026droid} & \cellcolor{secondc}0.3 & \cellcolor{secondc}0.3 & \cellcolor{secondc}0.2 & \cellcolor{secondc}0.3 & \cellcolor{secondc}0.7 & \cellcolor{secondc}0.5 & \cellcolor{secondc}0.4 & \cellcolor{thirdc}0.9 & \cellcolor{thirdc}2.7 & \cellcolor{firstc}\textbf{0.2} & \cellcolor{thirdc}0.65 \\
DROID-SLAM \cite{teed2021droid} & 0.6 & 4.7 & 1.2 & 2.3 & \cellcolor{firstc}\textbf{0.6} & 1.5 & 3.4 & 48.0 & 95.6 & 3.8 & 16.17 \\
MegaSaM \cite{li2025megasam} & 0.6 & 2.7 & 0.6 & 1.0 & 3.2 & 1.6 & 3.2 & 1.0 & 9.4 & \cellcolor{secondc}0.6 & 2.40 \\
\hdashline
\noalign{\vskip 1pt}
\multicolumn{12}{l}{\cellcolor[HTML]{EEEEEE}{\textit{Neural \& 3DGS SLAM}}} \\
NICE-SLAM \cite{zhu2022nice} & F & 123.6 & 21.1 & F & 150.2 & F & 134.4 & 138.4 & F & 23.8 & F \\
Splat-SLAM \cite{Sandstrom_2025_CVPR} & \cellcolor{thirdc}0.4 & \cellcolor{thirdc}0.4 & \cellcolor{thirdc}0.3 & \cellcolor{thirdc}0.7 & \cellcolor{thirdc}0.8 & \cellcolor{thirdc}0.6 & 1.9 & 2.5 & 73.6 & 5.9 & 8.71 \\
UP-SLAM~\cite{zheng2026upslam} & \cellcolor{thirdc}0.4 & 0.6 & 0.6 & 1.1 & 1.1 & 0.9 & \cellcolor{thirdc}1.0 & \cellcolor{secondc}0.7 & 3.6 & \cellcolor{thirdc}0.8 & 1.08 \\
WildGS-SLAM \cite{zheng2025wildgs} & \cellcolor{firstc}\textbf{0.2} & \cellcolor{firstc}\textbf{0.2} & \cellcolor{secondc}0.2 & \cellcolor{secondc}0.3 & \cellcolor{thirdc}0.8 & \cellcolor{firstc}\textbf{0.4} & \cellcolor{secondc}0.4 & \cellcolor{firstc}\textbf{0.6} & \cellcolor{firstc}\textbf{1.3} & \cellcolor{firstc}\textbf{0.2} & \cellcolor{firstc}\textbf{0.46} \\
\hdashline
SCOUT-SLAM(\textbf{Ours}) & \cellcolor{firstc}\textbf{0.2} & \cellcolor{firstc}\textbf{0.2} & \cellcolor{firstc}\textbf{0.1} & \cellcolor{firstc}\textbf{0.2} & \cellcolor{secondc}0.7 & \cellcolor{firstc}\textbf{0.4} & \cellcolor{firstc}\textbf{0.3} & \cellcolor{secondc}0.7 & \cellcolor{secondc}1.6 & \cellcolor{firstc}\textbf{0.2} & \cellcolor{secondc}0.47 \\
\bottomrule
\end{tabular}
}
\vspace{-2mm}
\caption{\textbf{Tracking Results on Wild-SLAM MoCap.} ATE RMSE [cm]. Best results are highlighted as \colorbox{firstc}{\textbf{first}}, \colorbox{secondc}{second}, and \colorbox{thirdc}{third}. 'F' indicated tracking failure. '-' denotes unreported sequences.}
\label{tab:mocap_results}
\vspace{-5pt}
\end{table*}
\subsection{Quantitative Results}
\subsubsection{Camera Tracking}
\cref{tab:bonn_results}, \cref{tab:tum_results}, and \cref{tab:mocap_results} show SCOUT-SLAM consistently achieves state-of-the-art performance across all three benchmarks. On TUM RGB-D~\cite{sturm12iros} (\cref{tab:tum_results}), SCOUT-SLAM attains an average error of 1.42~cm (vs.\ 1.51~cm for WildGS-SLAM~\cite{zheng2025wildgs} and 1.52~cm for DROID-W~\cite{li2026droid}). The largest gains occur on walking sequences (e.g., \textit{fr3/wr}: 2.91 vs.\ 3.15~cm) where dynamic objects dominate view. Baselines like DROID-SLAM~\cite{teed2021droid} and DynaMoN~\cite{schischka2024dynamon} perform favorably on \textit{fr2/dp}, as this sequence features a single actor remaining stationary for most of the scene without intermittent occlusions, allowing standard feature-matching and segmentation models to perform reliably. On Bonn Dynamic~\cite{refusion3d} (\cref{tab:bonn_results}), SCOUT-SLAM averages 2.29~cm compared to WildGS-SLAM~\cite{zheng2025wildgs} (2.47~cm) and DROID-W~\cite{li2026droid} (2.32~cm), with the largest margins appearing on \textit{balloon}, \textit{balloon2}, and \textit{crowd2} which contain rapidly moving objects. Finally, on Wild-SLAM MoCap (\cref{tab:mocap_results}), SCOUT-SLAM averages 0.47~cm, competitive with WildGS-SLAM~\cite{zheng2025wildgs} (0.46~cm) and outperforming DROID-W~\cite{li2026droid} (0.65~cm). Gains are most pronounced on sequences with large foreground motion (\textit{ANYmal2}, \textit{basketball}, \textit{crowd}, \textit{stones}). Performance on \textit{table\_tracking1}, \textit{table\_tracking2} is slightly lower than WildGS-SLAM~\cite{zheng2025wildgs}, likely due to a slow-moving foreground object near image boundaries producing an ambiguous MFC signal. This performance drop is more severe in DROID-W~\cite{li2026droid} due to the absence of reconstruction priors.
\subsubsection{Input View Synthesis TUM RGB-D~\cite{sturm12iros}} 
\cref{tab:tum_mapping} reports input view synthesis quality on the TUM RGB-D dynamic sequences. SCOUT-SLAM consistently improves PSNR, SSIM, and LPIPS over WildGS-SLAM~\cite{zheng2025wildgs} across all nine sequences (avg.\ PSNR: 19.34 vs.\ 18.91~dB), demonstrating the spatial-adaptive prior enforces complex static geometry reconstruction under fast camera motion(see \cref{fig:photo_cir_dep}).
\begin{table}[!t]
    \centering
    \footnotesize
    \renewcommand{\arraystretch}{0.85}
    \setlength{\tabcolsep}{4pt}
    \resizebox{\columnwidth}{!}{
    \begin{tabular}{lcc|lcc}
        \toprule
        \multicolumn{3}{c|}{\textbf{(a) Tracking Arch. (Bonn)}} & \multicolumn{3}{c}{\textbf{(b) Adaptive Prior (TUM)}} \\
        \midrule
        Architecture & Rank & ATE$\downarrow$ & Method & ATE$\downarrow$ & PSNR$\uparrow$\\
        \midrule
        TUA (Ours) & $r{=}4\,/\,8\,/\,16$ & 2.35\,/\,\textbf{2.29}\,/\,2.37 & Ours & \textbf{1.42} & \textbf{19.34} \\
        Standalone MLP & -- & 5.36 & w/o Prior & 1.45 & 19.20 \\
        \bottomrule
    \end{tabular}}
    \vspace{-2mm}
    \caption{Ablation studies.}
    \label{tab:ablations}
    \vspace{-2.5em}
\end{table}
\subsection{Qualitative Analysis}
\cref{fig:uncer_comparison} compares SCOUT-SLAM's tracking uncertainty estimation against WildGS-SLAM~\cite{zheng2025wildgs}, DROID-W~\cite{li2026droid} and MegaSAM~\cite{li2025megasam}. The results show that our method isolates dynamic objects precisely without affecting the static background. As shown in \cref{fig:photo_cir_dep}, WildGS-SLAM~\cite{zheng2025wildgs} assigns erroneously high mapping uncertainty to static regions under challenging lighting, complex geometry, and fast motion. This inflated uncertainty suppresses valid gradient updates and prevents the Gaussians from accurately reconstructing static surfaces, resulting in severe blurring and floaters. Whereas, SCOUT-SLAM yields sharp and artifact-free reconstructions under the same conditions. Furthermore, \cref{fig:uncer_convergence} illustrates tracking uncertainty convergence across BA iterations. While DROID-W~\cite{li2026droid} optimizes an affine layer using the MFC loss from scratch it requires more iterations to localize dynamic objects, whereas SCOUT-SLAM initializes its low-rank adapter (TUA) from the shared MUNet base weights. This pre-encoded scene representation allows the TUA to localize dynamic objects in fewer iterations while remaining stable thereafter as seen in \cref{fig:uncer_convergence}
\subsection{Ablation Studies} 
\cref{tab:ablations} presents our ablation study on TUM RGB Dynamic~\cite{sturm12iros}. (a) A standalone MLP of similar parameter count as low-rank matrices struggles to converge within the local BA window, with TUA rank=8 performing the best. (b) The network overestimates uncertainty on static regions, reducing both mapping and tracking accuracy. This shows that keeping the base network stable helps the tracking adapter converge.

\section{Conclusion}
\label{sec:conclusion}
We presented SCOUT-SLAM, a structurally-coupled dual-uncertainty framework that resolves the circular dependency between camera tracking and map reconstruction in 3DGS-SLAM. By deriving tracking and mapping uncertainties from a shared base network, our approach decouples tracking uncertainty from current reconstruction quality. Concurrently, a spatially-adaptive prior ensures stable uncertainty modeling and maintains stable features across the shared base network. Evaluations across dynamic benchmarks demonstrate that breaking this dependency enables SCOUT-SLAM to achieve state-of-the-art tracking accuracy and artifact-free static reconstructions in highly cluttered, dynamic environments.

{
    \small
    \bibliographystyle{ieeenat_fullname}
    \bibliography{main}
}
\newpage
\setcounter{section}{0}
\renewcommand\thesection{\Alph{section}}
\twocolumn[{%
\begin{center}
    \Large \textbf{Supplementary Material: \\ SCOUT-SLAM: Structurally-Coupled Dual Uncertainty-Aware 3DGS SLAM in the Wild}
    \vspace{0.8cm}
\end{center}
}]

This supplementary document provides additional details and extended experiments to support the main paper. Section~\ref{sec:eval} presents supplementary evaluations on the DROID-W dataset, details the datasets utilized, and outlines our extended evaluation masking protocol. Section~\ref{sec:impl} provides comprehensive implementation details, including the network architectures, the spatially-adaptive prior, and hyperparameters for the full SLAM pipeline. Finally, Section~\ref{sec:results} provides an extended runtime and memory analysis.

\section{Evaluation}\label{sec:eval}

\subsection{Evaluation on DROID-W}
\begin{figure}[ht]
    \centering
    \includegraphics[width=0.49\linewidth]{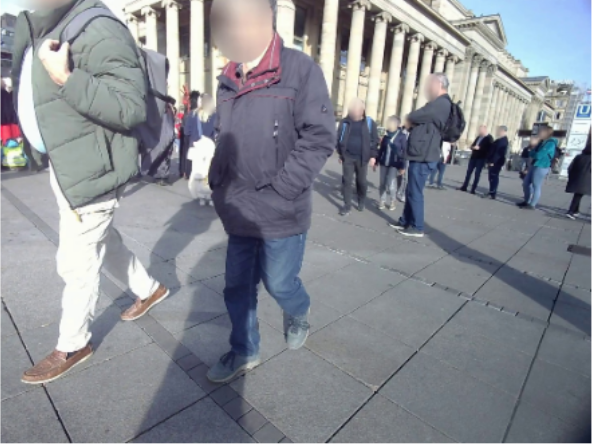} \hfill
    \includegraphics[width=0.49\linewidth]{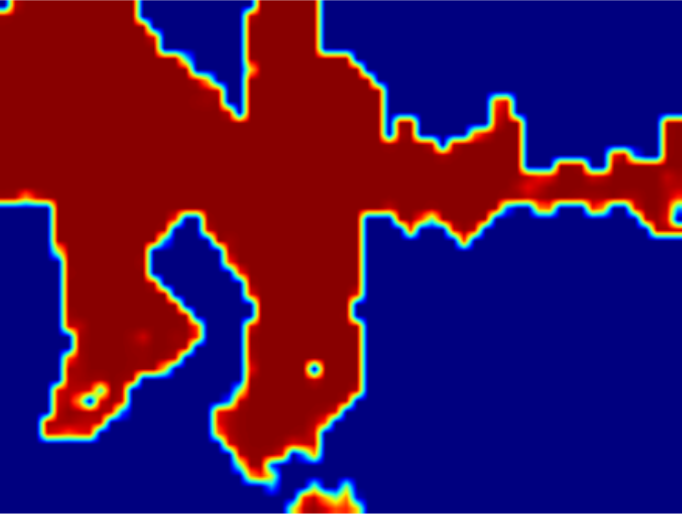} \\
    \vspace{1mm}
    \includegraphics[width=0.49\linewidth]{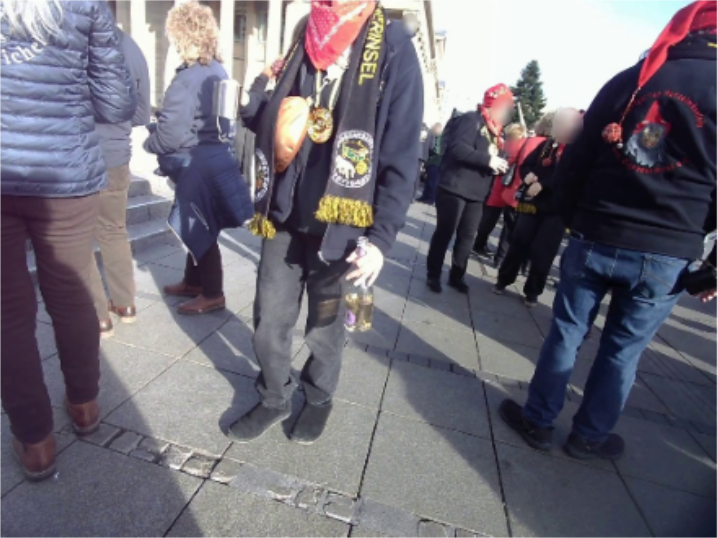} \hfill
    \includegraphics[width=0.49\linewidth]{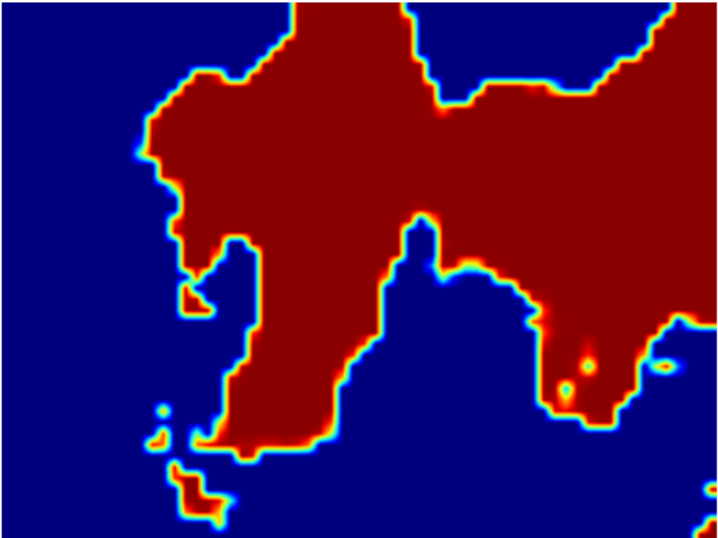} \\
    \vspace{1mm}
    \includegraphics[width=0.49\linewidth]{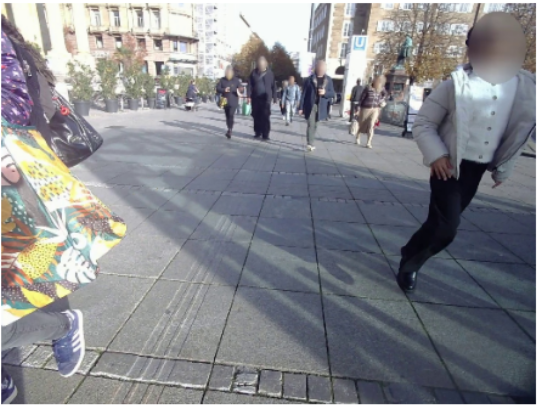} \hfill
    \includegraphics[width=0.49\linewidth]{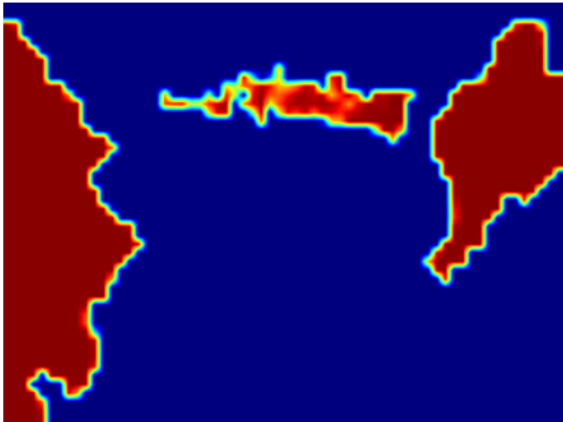}
    \caption{Qualitative visualization of tracking uncertainty in urban, cluttered dynamic environments. Left: Input RGB images. Right: Tracking uncertainty masks estimated by SCOUT-SLAM.}
    \label{fig:tracking_uncer}
\end{figure}

We extend our evaluation to the DROID-W~\cite{li2026droid} dataset (refer \cref{tab:droid_results}), which features urban outdoor environments with severe visual clutter and dynamic objects~\cref{fig:tracking_uncer}. In such challenging scenarios, the circular dependency between camera tracking and scene reconstruction is exacerbated, leading to poor uncertainty modeling, which our method addresses. Consequently, we compare our approach against WildGS-SLAM~\cite{zheng2025wildgs} and Splat-SLAM~\cite{Sandstrom_2025_CVPR}, to state-of-art 3DGS-SLAM methods. \texttt{Downtown 1} , \texttt{Downtown 4} sequences face Out-Of-Memory (OOM) issues owing to long unconstrained trajectories spanning more than 100 m, hence not reported. Crucially, the model architecture, hyperparameters, and training regime used for this new dataset are exactly identical to those evaluated in the main paper.
\begin{table*}[t]
\centering
\footnotesize
\begin{tabular}{lcccccc}
\toprule
Method & \texttt{Downtown 2} & \texttt{Downtown3} & \texttt{Downtown5} & \texttt{Downtown 6} & \texttt{Downtown7} & Avg. \\
\midrule
\scriptsize Splat-SLAM~\cite{Sandstrom_2025_CVPR} & \cellcolor{thirdc}6.44 & \cellcolor{thirdc}0.89 & \cellcolor{thirdc}0.91 & \cellcolor{thirdc}2.11 & \cellcolor{firstc}\textbf{0.07} & \cellcolor{thirdc}2.08 \\
\scriptsize WildGS-SLAM \cite{zheng2025wildgs} & \cellcolor{secondc}0.95 & \cellcolor{secondc}0.43 & \cellcolor{secondc}0.87 & \cellcolor{secondc}1.22 & \cellcolor{secondc}0.53 & \cellcolor{secondc}0.80 \\
\hdashline
\scriptsize SCOUT-SLAM(\textbf{Ours}) & \cellcolor{firstc}\textbf{0.79} & \cellcolor{firstc}\textbf{0.23} & \cellcolor{firstc}\textbf{0.86} & \cellcolor{firstc}\textbf{0.09} & \cellcolor{firstc}\textbf{0.07} & \cellcolor{firstc}\textbf{0.41} \\
\bottomrule
\end{tabular}
\caption{\textbf{Tracking Results on DROID-W\cite{li2026droid} Dataset Sequences.} ATE RMSE [m]. Best results are highlighted as \colorbox{firstc}{\textbf{first}}, \colorbox{secondc}{second}, and \colorbox{thirdc}{third}.}
\label{tab:droid_results}
\end{table*}

\subsection{Datasets}
We evaluate SCOUT-SLAM on dynamic sequences with rapid camera motion and complex lighting. To balance tracking stability and real-time mapping speed, sequences are processed at reduced resolutions:
\begin{itemize}
    \item \textbf{TUM RGB-D}: Features indoor dynamic interactions (e.g., people walking). Originally $480 \times 640$, processed at $384 \times 512$ to maintain a $4:3$ aspect ratio while meeting network divisibility constraints.
    \item \textbf{Bonn Dynamic}: Contains ground-truth poses under severe dynamic occlusions (e.g., moving persons, balloons, boxes). Processed at $384 \times 512$.
    \item \textbf{Wild-SLAM MoCap}: Features open-set dynamic objects (quadrupeds, basketballs, crowds) in uncontrolled environments. Processed at $360 \times 640$.
    \item \textbf{DROID-W}: Features urban outdoor environments with severe visual clutter and dynamic objects. Originally $1200 \times 1600$, processed at $384 \times 512$.
    \item \textbf{Youtube}: Unconstrained internet videos with aggressive camera motion and heavy visual clutter. Processed at $384 \times 512$.
\end{itemize}

\subsection{Input view synthesis evaluation.}
TUM RGB-D sequences lack ground-truth dynamic masks, which are necessary to evaluate the reconstruction quality of the static background geometry. We use SAM3 to generate dynamic object masks for both rendered and ground-truth images. Specifically, we manually segment and exclude the "person" and "chair" categories from the evaluation metrics (PSNR, SSIM, LPIPS) across all TUM sequences, ensuring that metrics reflect only static scene fidelity.

\begin{figure}[h]
    \centering
    \includegraphics[width=0.24\linewidth]{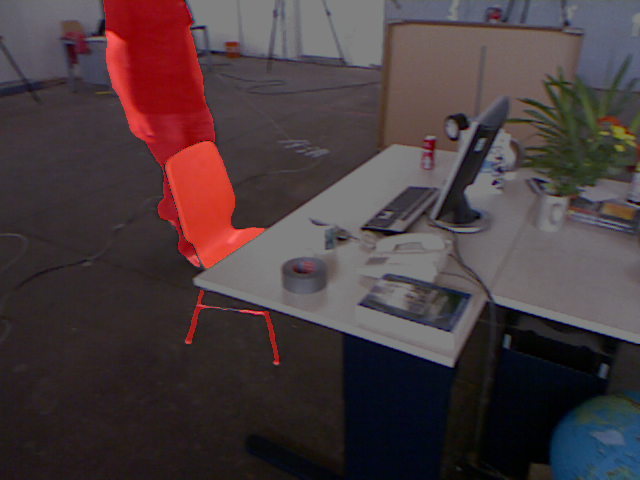} \hfill
    \includegraphics[width=0.24\linewidth]{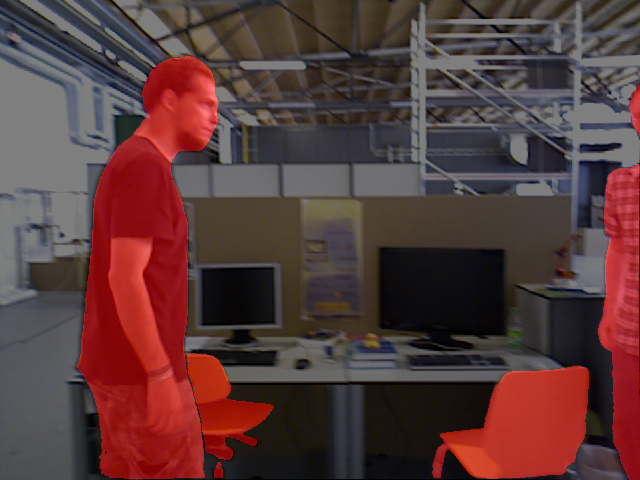} \hfill
    \includegraphics[width=0.24\linewidth]{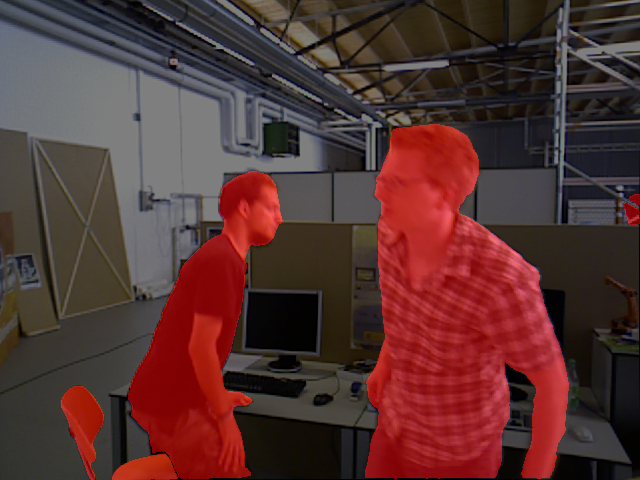} \hfill
    \includegraphics[width=0.24\linewidth]{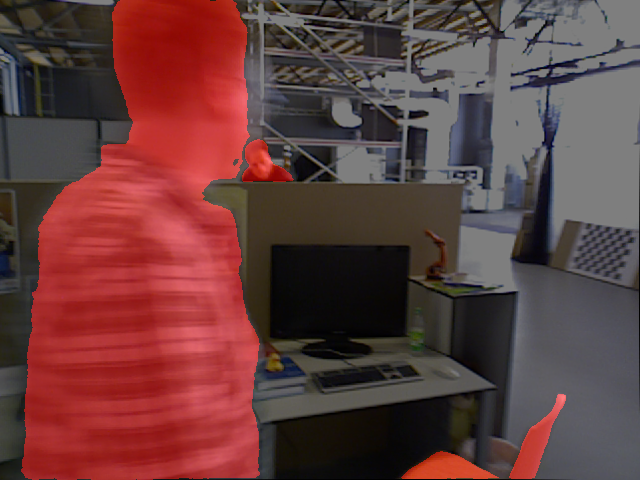}
    \caption{Examples of dynamic object masks overlayed on ground-truth images generated by SAM3 for evaluation.}
    \label{fig:eval_masks}
\end{figure}

\section{Implementation Details}\label{sec:impl}

\subsection{Network Architectures}
SCOUT-SLAM extracts multi-view features using a frozen DINOv2 (\texttt{vits14\_reg}) encoder, yielding a 384-dimensional feature map.
\begin{itemize}
    \item \textbf{Mapping Uncertainty Network (MUNet):} A 2-layer Multi-Layer Perceptron (MLP) with a hidden dimension of 64. It uses ReLU activations for hidden layers and a Softplus activation at the output to predict a strictly positive per-pixel mapping uncertainty.
    \item \textbf{Tracking Uncertainty Adapter (TUA):} TUA injects trainable low-rank matrices ($r=8$, scaling factor $\alpha=8.0$) into the MUNet layers. During tracking, MUNet's base weights are frozen. Only the low-rank matrices are optimized via AdamW (learning rate $4\times10^{-7}$, weight decay $0.05$).
\end{itemize}

\subsection{Spatially-Adaptive Prior}
The spatially-adaptive prior $\lambda(w, R)$ replaces the uniform penalty in the Negative Log-Likelihood loss. It is conditioned on the tracking confidence $w$ and the combined rendering residual $R$ (SSIM and depth error). 

It takes the form: $\lambda(w, R) = \lambda_0 + \gamma \cdot w \cdot g(R)$, where $g(R) = \exp(-\alpha R)$ is a residual gating mechanism. In our configuration, we set the base penalty $\lambda_0 = 0.98$, the prior scale $\gamma = 2.0$, and the gating parameter $\alpha = 0.6$. The combined residual is computed as $R = \mathcal{L}_{\text{SSIM}} + 0.2 \cdot \mathcal{L}_{\text{depth}}$. 
Primarily, $\lambda$ regularizes uncertainty inflation on static geometry (where confidence $w$ is high). However, if a static region exhibits  residual $R$ due to transient variations (e.g., sensor noise or lighting flicker), $g(R)$ approaches 0. This softens the penalty back to $\lambda_0$, allowing the network to safely assign uncertainty without forcing the 3D Gaussians to overfit to the transient noise.

\subsection{SLAM Pipeline \& Hyperparameters}
The system runs on an Intel Core i9 CPU and an NVIDIA RTX 4090 GPU. 

\paragraph{Multi-Processing Architecture.}
\label{para: mp}
To ensure real-time performance, SCOUT-SLAM decouples camera tracking and scene reconstruction into asynchronous parallel processes using \texttt{torch.multiprocessing}. The tracker estimates poses and dispatches keyframe data (poses, depth, flags) to the mapper via a unidirectional inter-process pipe. To prevent the mapper from being overwhelmed and maintain graph consistency, the processes synchronize using continuation signals dispatched by the mapper upon completing a keyframe optimization step. An optional third process handles live GUI rendering, receiving visualization data via asynchronous queues.

\paragraph{System Initialization.} 
SCOUT-SLAM uses a two-stage initialization. First, incoming frames are buffered until a 15-keyframe warmup window is reached. The tracking frontend establishes initial poses, and the mapper uses this geometry to seed the 3D Gaussians and train the MUNet for 1050 iterations. 
Second, the frontend updates the tracking uncertainty masks for these warmup frames using the trained MUNet via TUA. The initial tracking graph is then re-optimized using these uncertainty weights to suppress dynamic outliers before standard online SLAM begins.

\paragraph{Tracking Frontend.} 
New keyframes are added when the average optical flow magnitude exceeds 3.0 pixels. Poses are optimized within a 25-keyframe sliding bundle adjustment window, capped at 75 edges. Intermediate non-keyframe poses are linearly interpolated in $SE(3)$ log-space.

\paragraph{Global Bundle Adjustment.} 
To maintain long-term trajectory consistency, SCOUT-SLAM triggers an online Global Bundle Adjustment (BA) every 20 keyframes, as well as a final BA at the end of the sequence, similar to WildGS-SLAM~\cite{zheng2025wildgs}. During Global BA, the low-rank matrices of the Tracking Uncertainty Adapter (TUA) are not optimized. Instead, historical tracking uncertainty masks are held fixed while poses and metric depths are refined using extended loop closure factors. When global BA updates historical camera poses, the 3D Gaussians in the map are correspondingly deformed to align with the refined trajectory.

\paragraph{Dense Mapping.} 
Metric depth priors are generated using Metric3D~v2. For initializing 3D Gaussians, images and depth maps are spatially downsampled by a factor of $32$. To accelerate optimization and reduce memory overhead, we set the spherical harmonics (SH) degree of the 3D Gaussians to 0. Gaussians are optimized over a 10-keyframe co-visibility window for 450 iterations per new keyframe. We prune Gaussians with an opacity below 0.7 or a screen radius exceeding 20 pixels.

\section{Results}\label{sec:results}
\subsection{Runtime Analysis}
Table~\ref{tab:runtime} details keyframe latency and memory usage on TUM RGB-D sequences. SCOUT-SLAM achieves better tracking accuracy ($1.42$ vs.\ $1.51$~cm) with minimal overhead, maintaining comparable mapping speeds ($\sim$980 vs.\ $\sim$910~ms/kf) and a modest tracking increase ($\sim$710 vs.\ $\sim$470~ms/kf). Crucially, our custom CUDA implementation ensures the dense MFC loss adds negligible memory overhead. Thus, overall peak VRAM remains highly competitive at ($\sim$6.8~GB vs.\ $\sim$6.1~GB), preserving system efficiency while robustly constraining dynamic regions.
\begin{table}[t]
\centering
\footnotesize
\setlength{\tabcolsep}{4pt}
\resizebox{\linewidth}{!}{%
\begin{tabular}{lcccc}
\toprule
Method & Track (ms/kf) & Map (ms/kf) & Peak VRAM (GB) & ATE [cm] \\
\midrule
WildGS-SLAM~\cite{zheng2025wildgs} & ~\textbf{470} & ~\textbf{910} & \textbf{6.1} & 1.51 \\
SCOUT-SLAM (\textbf{Ours}) & ~710 & ~\textbf{980} & 6.8 & 1.\textbf{42} \\
\bottomrule
\end{tabular}%
}
\vspace{-2mm}
\caption{\textbf{Runtime Analysis} on TUM RGB-D dynamic sequences. Average latency per keyframe (ms/kf), peak VRAM, and ATE RMSE [cm] across the TUM sequences.}
\label{tab:runtime}
\vspace{-5pt}
\end{table}
\paragraph{Future Work.} 
Similar to existing 3DGS-SLAM methods, overall runtime is 
bottlenecked by the synchronous operation (refer~\cref{para: mp}) of tracking and mapping. Exploring asynchronous architectures that leverage the reduced tracking-mapping coupling is a promising direction for real-time deployment.

\end{document}